\documentclass[ba]{imsart}

\RequirePackage{amsthm,amsmath,amsfonts,amssymb}
\RequirePackage[numbers]{natbib}
\RequirePackage[colorlinks,citecolor=blue,urlcolor=blue,backref=page,backref=page]{hyperref}
\RequirePackage{graphicx}
\usepackage{bbm}
\usepackage{amsmath}
\usepackage{tikz}
\usepackage{subcaption} 
\usetikzlibrary{positioning, arrows.meta, backgrounds, fit, decorations.pathreplacing}

\newcommand{\plate}[4][black]%
{
  \begin{scope}[xshift=#2, yshift=#3, rotate=#4, color=#1, thick, rounded corners=2pt, dashed]
    \draw (-0.2,0.2) rectangle (0.2,-0.2) ;
    \draw (-0.2,-0.2) -- (-0.4,-0.4) ;
    \draw (0.2,-0.2) -- (0.4,-0.4) ;
  \end{scope}
}

\newcommand{\bx}{\mathbf{x}}

\newcommand{\bX}{\mathbf{X}}
\newcommand{\bY}{\mathbf{Y}}

\renewcommand{\theenumi}{\Alph{enumi}}
\newcommand{\bth}{\boldsymbol{\theta}}

\newcommand{\ben}{\begin{eqnarray*}}
\newcommand{\een}{\end{eqnarray*}}

\newcommand{\be}{\begin{eqnarray}}
\newcommand{\ee}{\end{eqnarray}}

\DeclareMathOperator*{\argmax}{arg\,max}
\DeclareMathOperator*{\argmin}{arg\,min}

\pubyear{2024}
\volume{TBA}
\issue{TBA}
\firstpage{1}
\lastpage{1}
\usepackage{comment}
\startlocaldefs
\theoremstyle{plain}

\newtheorem{theorem}{Theorem}[section]
\newtheorem{prop}{Propositions}[section]

\theoremstyle{definition}

\theoremstyle{remark}

\endlocaldefs

\begin{document}

\begin{frontmatter}
\title{Bayesian Nonparametrics: An Alternative to Deep Learning}
\runtitle{Bayesian Nonparametrics}

\begin{aug}
\author[A]{\fnms{Bahman}~\snm{Moraffah}\ead[label=e1]{Bahman.Moraffah@asu.edu}\orcid{0000-0002-9806-8948}}
\address[A]{School of Electrical Engineering,
Arizona State University\printead[presep={,\ }]{e1}}

\end{aug}

\begin{abstract}
Bayesian nonparametric models offer a flexible and powerful framework for statistical model selection, enabling the adaptation of model complexity to the intricacies of diverse datasets. This survey intends to delve into the significance of Bayesian nonparametrics, particularly in addressing complex challenges across various domains such as statistics, computer science, and electrical engineering.
By elucidating the basic properties and theoretical foundations of these nonparametric models, this survey aims to provide a comprehensive understanding of Bayesian nonparametrics and their relevance in addressing complex problems, particularly in the domain of multi-object tracking. Through this exploration, we uncover the versatility and efficacy of Bayesian nonparametric methodologies, paving the way for innovative solutions to intricate challenges across diverse disciplines.
\end{abstract}



\end{frontmatter}

\maketitle
\tableofcontents

\section{Introduction}

In recent years, deep learning has emerged as the de facto standard for tackling a myriad of complex tasks in artificial intelligence, ranging from image recognition to natural language processing. Its remarkable success can be attributed to its ability to learn intricate patterns and representations from vast amounts of data. However, as the field matures, it becomes increasingly evident that deep learning is not a one-size-fits-all solution, especially when faced with certain challenges such as data scarcity, uncertainty quantification, and interpretability.

In this context, Bayesian nonparametrics offers a compelling alternative approach that complements and, in some cases, surpasses the capabilities of deep learning. While deep learning relies on fixed-dimensional parameter spaces and often requires large amounts of labeled data for training, Bayesian nonparametrics provides a flexible framework that can adapt to the complexity of the data and incorporate uncertainty in a principled manner.

Deep learning architectures, while powerful, often face constraints due to their fixed number of parameters predetermined before training, which can limit their adaptability to the complexity inherent in diverse datasets. In contrast, Bayesian nonparametric models offer a flexible solution by allowing for an infinite-dimensional parameter space. This flexibility enables them to capture intricate patterns and structures within the data without the need for manual specification of model complexity, thereby accommodating the diverse and nuanced nature of real-world data.

One significant drawback of deep learning models is their limited ability to quantify uncertainty, which is essential for making informed decisions in critical applications such as medical diagnosis and financial forecasting. Bayesian nonparametric methods address this limitation by inherently incorporating uncertainty through the generation of posterior distributions over model parameters. This integration of uncertainty enables more robust decision-making in environments where uncertainty is prevalent, enhancing the reliability and trustworthiness of model predictions.

Furthermore, while deep learning thrives in tasks with ample labeled data and computational resources, it often struggles when faced with limited data or computational constraints. In contrast, Bayesian nonparametric methods demonstrate resilience in such scenarios. By leveraging the inherent structure of the data and borrowing strength across observations, these methods can effectively extract meaningful insights from small datasets. Additionally, Bayesian nonparametric models employ efficient inference algorithms that facilitate scalability to large datasets, ensuring their applicability across a wide range of data-intensive tasks while maintaining computational efficiency.

This survey is designed to serve as the cornerstone of exploration, fostering critical insights that are indispensable for propelling progress in a multitude of domains. Whether in healthcare, finance, climate science, object tracking, or beyond, its findings hold the potential to revolutionize existing paradigms and unlock new avenues of inquiry \cite{moraffah2019bayesian, teh2006hierarchical, venkataraman2001statistical, sharif2008overview, mackay1995hierarchical}. By embracing a holistic approach, this survey not only illuminates current challenges but also paves the way for innovative solutions, positioning itself as a versatile tool capable of complementing or even surpassing the capabilities of deep learning in certain contexts.

 Bayesian nonparametric models offer versatile statistical model selection techniques, facilitating the choice of an appropriate level of complexity across a spectrum of challenges in statistics, computer science, and electrical engineering. With a primary focus on addressing issues arising in multi-object tracking, this thesis endeavors to harness the potential of Bayesian nonparametric models in tackling such complex problems.

To lay the groundwork, we commence with a comprehensive examination of key distributions pivotal in Bayesian statistics, emphasizing the significance of conjugate priors in Bayesian analysis. Subsequently, we delve into the fundamental principles of Bayesian nonparametrics, elucidating their importance and applicability in diverse problem domains including in deep neural networks \cite{chen2011hierarchical, guo2018explaining}. This survey does not encompass a class of Bayesian nonparametrics known as diffusion trees. Interested readers can find more information in \cite{neal2003density, neal2006high, heaukulani2014beta, knowles2014pitman,Moraffah2019Tree, knowles2011pitman}.

In subsequent sections, we embark on a detailed exploration of prominent nonparametric models that underpin the development of novel methodologies in this thesis. We commence with an in-depth discussion of the Dirichlet process and its properties, followed by an exploration of the generalized Dirichlet process. Furthermore, we conduct a thorough analysis of the Two-parameter Poisson Dirichlet process (commonly known as the Pitman-Yor Process), elucidating its intricate properties and implications for Bayesian inference.

\section{Background}

The adaptation of Bayesian inferential methods to accommodate the unique characteristics of nonparametric models is imperative for meaningful analysis. The advent of Markov chain Monte Carlo (MCMC) methods has revolutionized Bayesian inference, enabling robust analysis in high-dimensional datasets,  \cite{robert2013monte}. In this context, we explore two primary inference methods, namely Monte Carlo methods and variational Bayes inference methods, which form the cornerstone of flexible and robust analysis in infinite-dimensional spaces. Building upon these methods, we propose novel inferential models tailored to provide tractable analysis of Bayesian nonparametric models, thus unlocking new avenues for addressing complex challenges across diverse domains.

\section{Exploring Parametric Distributions}
\subsection{Understanding the Exponential Family}
\label{exp}

In this section, we delve into a versatile class of parametric distributions known as the exponential family. This family encompasses a wide range of distributions, including but not limited to Gaussian, multinomial, Poisson, and Beta distributions. Consider a random variable $\bX \in \mathcal{X}$ and a distribution family characterized by the density function (given $\bth$):

\begin{equation}
\label{expfam}
p(\bx|\bth) = h(\bx)\exp\{\bth^T T(\bx) - A(\bth)\}.
\end{equation}

Here, $\bth$ represents the family's natural or canonical parameters, $h(\bx)$ is a nonnegative reference measure, and $T(\bx)$ denotes the sufficient statistics. The cumulant function $A(\bth)$, defined as the logarithm of a normalizer, ensures normalization of the distribution:

\begin{equation}
\label{cumu}
A(\bth) = \log \int h(\bx) \exp\{\bth^T T(\bx)\} \nu(d\bx),
\end{equation}

where $\nu$ is a measure. The exponential family is well-defined if the integral in \ref{cumu} is finite. The set of canonical parameters for which \ref{cumu} is finite is termed the natural parameter space ($\mathcal{C}$) and is defined as:

\begin{equation}
\mathcal{C} = \{\bth : A(\bth) < \infty\}.
\end{equation}

We focus on regular exponential families, where $\mathcal{C}$ is a nonempty open set. Notably, distributions like Gaussian, Poisson, Beta, and Gamma fall within this category. The convexity of $A(\bth)$ in $\bth$ implies the convexity of $\mathcal{C}$, and in minimal families, $A(\bth)$ is strictly convex \cite{wainwright2008graphical}. Additionally, there exists a profound relationship between the derivatives of the cumulant function and the moments of sufficient statistics \cite{brown1986fundamentals, wainwright2008graphical, robert2007bayesian}:

\begin{equation}
\begin{split}
\frac{\partial A}{\partial \bth^T} &= \mathbb{E}[T(\bX)],\\
\frac{\partial^2 A}{\partial \bth \partial \bth^T} &= \text{Var}[T(\bX)].
\end{split}
\end{equation}

\subsubsection{Maximum Likelihood Estimator}
In this section, we explore the maximum likelihood estimator ($\hat{\mu} := \mathbb{E}[T(\bX)]$) as a function of the canonical parameter $\bth$. Assuming $\bx_1, \dots, \bx_N \sim p(\bx|\bth)$, the log-likelihood is given by:

\begin{equation}
\ell(\bth) = \log \left(\prod\limits_{j=1}^{N} h(\bx_j)\right) + \bth^T \left(\sum\limits_{j=1}^{N} T(\bx_j)\right) - N A(\bth).
\end{equation}

Setting the partial derivative to zero yields the unbiased maximum likelihood estimator of $\bth$:

\begin{equation}
\hat{\bth}_\text{MLE} = \frac{1}{N} \sum\limits_{j=1}^{N} T(\bx_j).
\end{equation}

This estimator is unbiased and attains the Cramér-Rao lower bound, where the Fisher information is inversely proportional to the variance of $T(\bX)$:

\begin{equation}
\text{Fisher Information} = \frac{1}{\text{Var}[T(\bX)]}.
\end{equation}

Considering i.i.d. samples $\bx_1, \dots, \bx_N \sim \tilde{p}$, the empirical density $p^*(\bx)$ of the samples is:

\begin{equation}
p^*(\bx) = \frac{1}{N} \sum \limits_{j=1}^{N} \delta_{\bx_j}(\bx),
\end{equation}

where $\delta_{\bx_j}(\bx)$ denotes the Dirac delta function. Remarkably, there exists a significant connection between maximizing the likelihood and minimizing the Kullback-Leibler (KL) divergence, offering an intriguing alternative approach to optimize the likelihood function.

The equation below, known as the Kullback-Leibler divergence, quantifies the difference between two probability distributions $p^*$ and $p_{\bth}$:

\begin{equation}
\label{klml}
\begin{split}
KL(p^*, p_{\bth}) &= \sum\limits_\bx p^*(\bx) \log \frac{p^*(\bx)}{p(\bx|\bth)}  \\
& = \sum\limits_\bx p^*(\bx) \log p^*(\bx) - \sum\limits_\bx p^*(\bx) \log p(\bx|\bth) \\
& = - H(p^*) - \sum\limits_\bx \frac{1}{N} \sum \limits_{j=1}^{N} \delta_{\bx_j}(\bx) \log p(\bx|\bth)\\
& =   - H(p^*) - \frac{1}{N} \sum \limits_{j=1}^{N}  \log p(\bx_j|\bth)  \\
& = - H(p^*) - \frac{1}{N} \ell(\bth)
\end{split}
\end{equation}

Here, $H(p^*)$ denotes the entropy of $X$ with respect to the empirical density $p^*$, a quantity independent of $\bth$. Equation \ref{klml} highlights the relationship:

\begin{equation}
\hat{\bth}_{\text{MLE}} = \argmax_{\bth} \ell(\bth) = \argmin_{\bth} KL(p^*, p_{\bth}).
\end{equation}

Thus, the maximum likelihood estimator $\hat{\bth}_{\text{MLE}}$ can be effectively obtained by minimizing the Kullback-Leibler divergence between the empirical density $p^*$ and the distribution $p_{\bth}$. This observation provides valuable insight into optimizing likelihood functions through a different lens.

\subsubsection{Bayesian Inference}

Up to this point, we've considered parameters as fixed but unknown. Now, we delve into Bayesian inference, treating parameters as random variables. A detailed exploration of this topic is available in references such as \cite{bernardo2007bayesian, robert2007bayesian}.

Let's assume $\bx_1, \dots, \bx_N \sim p(\bx|\bth)$, where $p(\bx|\bth)$ represents the canonical exponential family with parameter $\bth$. We introduce a prior $p(\bth|\gamma)$ on the parameter $\bth$ with hyperparameters $\gamma$. By Bayes' rule, the posterior distribution is given by:

\begin{equation}
p(\bth|\{\bx_j\}_{j=1}^{N}, \gamma) \propto  p(\bth|\gamma) \prod\limits_{j=1}^{N} p(\bx_j|\bth).
\end{equation} 

In Bayesian statistics, we typically assign a prior $p(\gamma)$ over hyperparameters $\gamma$. However, estimating $\gamma$ often involves frequentist methods, such as:

\begin{equation}
\hat{\gamma} = \argmax_{\gamma} p(\bx_1, \dots, \bx_N | \gamma).
\end{equation}

However, this optimization problem is often intractable. The best $\gamma$ can be approximated using techniques like leave-one-out cross-validation. The predictive distribution is given by:

\begin{equation}
p(\bx_\text{new}|\{\bx_j\}_{j=1}^{N}, \gamma) = \int_{\mathcal{C}} p(\bx_\text{new}|\bth) p(\bth|\{\bx_j\}_{j=1}^{N}, \gamma) d\bth,
\end{equation}

which can be intractable in some cases. In such instances, we resort to approximating parameters using the MAP estimator:

\begin{equation}
\hat{\bth}_\text{MAP} = \argmax_{\bth} p(\bth|\{\bx_j\}_{j=1}^{N}, \gamma) .
\end{equation}

We focus on cases where the predictive distribution is tractable, termed as conjugate priors.

\textbf{Definition:} A family of distributions, $\mathcal{F}$, is termed a conjugate prior for likelihood $p(\bx|\bth)$ if for every prior $p(\bth)\in \mathcal{F}$, the posterior $p(\bth|\bx) \in \mathcal{F}$.

For the exponential family with density given by \ref{expfam}, the likelihood for independent samples $\bx_1, \dots, \bx_N$ is:

\begin{equation}
p(\bx_1,\dots,\bx_N|\bth) = \bigg(\prod\limits_{j=1}^{N} h(\bx_j)\bigg) \exp\bigg(\bth^T\bigg(\sum\limits_{j=1}^{N} T(\bx_j)\bigg) - N A(\bth) \bigg).
\end{equation}

\textbf{Proposition:} A conjugate family for the likelihood $p(\bx_1,\dots,\bx_N|\bth)$ is given by:

\begin{equation}
p(\bth|\tau, \zeta) = Z(\tau, \zeta) \exp\bigg(\tau^T\bth - \zeta A(\bth)\bigg),
\end{equation}

where $Z(\tau, \zeta)$ is the normalizer. Then, the posterior distribution is:

\begin{equation}
p(\bth|\tau+\sum\limits_{j=1}^N T(\bx_j), \zeta + N).
\end{equation}

With Proposition 4 in mind, we can compute the predictive likelihood as:

\begin{equation}
p(\bx_\text{new}|\{\bx_j\}_{j=1}^{N}, \gamma) = \frac{Z(\tau+\sum\limits_{j=1}^N T(\bx_j),\zeta + N) }{Z(\tau+\sum\limits_{j=1}^N T(\bx_j)+T(\bx_\text{new}),\zeta + N+1)}.
\end{equation}

For further details and proof, refer to \cite{robert2007bayesian,sudderth2006graphical}. Additionally, it can be shown that the posterior expectation of $\mu = \mathbb{E}[T(\bX)]$ is a convex combination of the prior expectation and the maximum likelihood estimate.

\section{Foundational Probability Distributions}

\subsection{Multinomial Distribution:}

Consider a random variable $\bX$ with $K$ categorical values, where each value has a probability $\pi_k = \mathbb{P}(\bX = k)$. The distribution of $\{\bX_k\}_{k=1}^K$ follows a multinomial distribution, denoted by Multi$(n; p_1,\dots, p_K)$ \cite{gelman2013bayesian, robert2007bayesian}, with the probability mass function:

\begin{equation}
p(\bx_1, \dots, \bx_K) = \Bigg(\frac{n!}{\prod_k\bx_k!}\prod\limits_{k=1}^K \pi^{\bx_k}_k \Bigg)\mathbbm{1}_{[\sum_k\bx_k]}(n).
\end{equation}

The parameters of a multinomial distribution lie in a $K-1$ dimensional simplex:

\begin{equation}
\Pi_{K-1} = \{\pi\in\mathbb{R}^K: 0 \leq \pi_k \leq1, \sum_k \pi_k = 1\}.
\end{equation}

The mean and variance of $\bX_k$ are $\mathbb{E}[\bX_k] = n\pi_k$ and $\text{Var}(\bX_k) = n\pi_k(1-\pi_k)$, respectively.

The multinomial distribution can be re-expressed as:

\begin{equation}
p(\bx_1, \dots, \bx_K) = \Bigg(\frac{n!}{\prod_k\bx_k!}\exp\bigg( \sum\limits_{k=1}^K \bx_k \log \pi_k\bigg) \Bigg)\mathbbm{1}_{[\sum_k\bx_k]}(n).
\end{equation}

Thus, it defines a regular exponential family with canonical parameters $\bth_k = \log\big(\frac{\pi_k}{1-\sum_{k=1}^{K-1} \pi_k}\big)$ and cumulant $A(\bth) = -\log\big(1-\sum_{k=1}^{K-1} \pi_k\big)$. The maximum likelihood estimator of the multinomial parameters is $\hat{\pi}_k = \bx_k/n$.

\subsection{Dirichlet Distribution:}

The Dirichlet distribution, a fundamental concept in Bayesian statistics and probabilistic modeling, serves as the conjugate prior for the multinomial distribution, providing a flexible framework for modeling categorical data \cite{robert2007bayesian, sudderth2006graphical}. Denoted as $\text{Dir}(\alpha_1, \dots,\alpha_K)$, where $(\alpha_1, \dots,\alpha_K)$ are its parameters, this distribution encompasses a range of probability distributions.

In its essence, the Dirichlet distribution captures the uncertainty over the probabilities of observing different outcomes in a categorical event. Its probability density function, given by:

\begin{equation}
p(\pi_1,\dots,\pi_K|\boldsymbol{\alpha}) = \frac{\Gamma(\sum_{k=1}^K \alpha_k)}{\prod_{k=1}^K \Gamma(\alpha_k)}\prod_{k=1}^K \pi_k^{\alpha_k-1}
\end{equation}

Where $\pi = (\pi_1,\dots,\pi_K) \in \Pi_{K-1}$, represents a point in the K-1 dimensional simplex, ensuring that the probabilities sum up to one.

For $K=2$, the Dirichlet distribution reduces to the Beta distribution, a widely used distribution in modeling binary events. Further elaboration on the Beta distribution will be provided in the subsequent section.

The properties of the Dirichlet distribution offer valuable insights into its behavior and utility:

\begin{prop}
If $\boldsymbol{\pi} \sim \text{Dir}(\alpha_1, \dots, \alpha_K)$, and $\alpha_0 = \sum_k \alpha_k$, then:
\begin{enumerate}
\setlength\itemsep{0.3em}
\item[a)] The expected value of each component $\pi_k$ is given by $\mathbb{E}[\pi_k] = \frac{\alpha_k}{\alpha_0}$.
\item[b)] The variance of $\pi_k$ is $\text{Var}(\pi_k) = \frac{\alpha_k (\alpha_0 - \alpha_k)}{\alpha^2_0(\alpha_0 +1)}$.
\item[c)] The covariance between $\pi_j$ and $\pi_k$ is $\text{Cov}(\pi_j,\pi_k) = \frac{-\alpha_j\alpha_k}{\alpha^2_0(\alpha_0 +1)}$, where $j\neq k$.
\item[d)] \textbf{Aggregation property:} Combining a subset of categories remains Dirichlet. For example, if $\boldsymbol{\pi} \sim \text{Dir}(\alpha_1, \dots,\alpha_{K-1}, \alpha_K)$, combining the last two components yields $(\pi_1,\dots, \pi_{K-1}+\pi_K) \sim \text{Dir}(\alpha_1, \dots,\alpha_{K-1}+\alpha_K)$.
\item[e)] The marginal distribution of any individual $\pi_k$ follows a Beta density, i.e., $\pi_k \sim \text{Beta}(\alpha_k, \alpha_0 - \alpha_k)$.
\item [f)] Multinomial and Dirichlet distributions exhibit conjugacy. The posterior distribution after observing $N$ events $\{\bx^n\}_{n=1}^N$ from a multinomial distribution is Dirichlet-distributed, given by:
\begin{equation}
p(\boldsymbol{\pi}| \{\bx^k\}_{n=1}^N, \boldsymbol{\alpha}) \sim \text{Dir}(\alpha_1+\sum_n \mathbbm{1}_{\bx^n}(1), \dots, \alpha_K+\sum_n \mathbbm{1}_{\bx^n}(K)).
\end{equation}
\end{enumerate}
\end{prop} 
\begin{figure}[!t]
\centering
\includegraphics[width=9cm, height = 6cm]{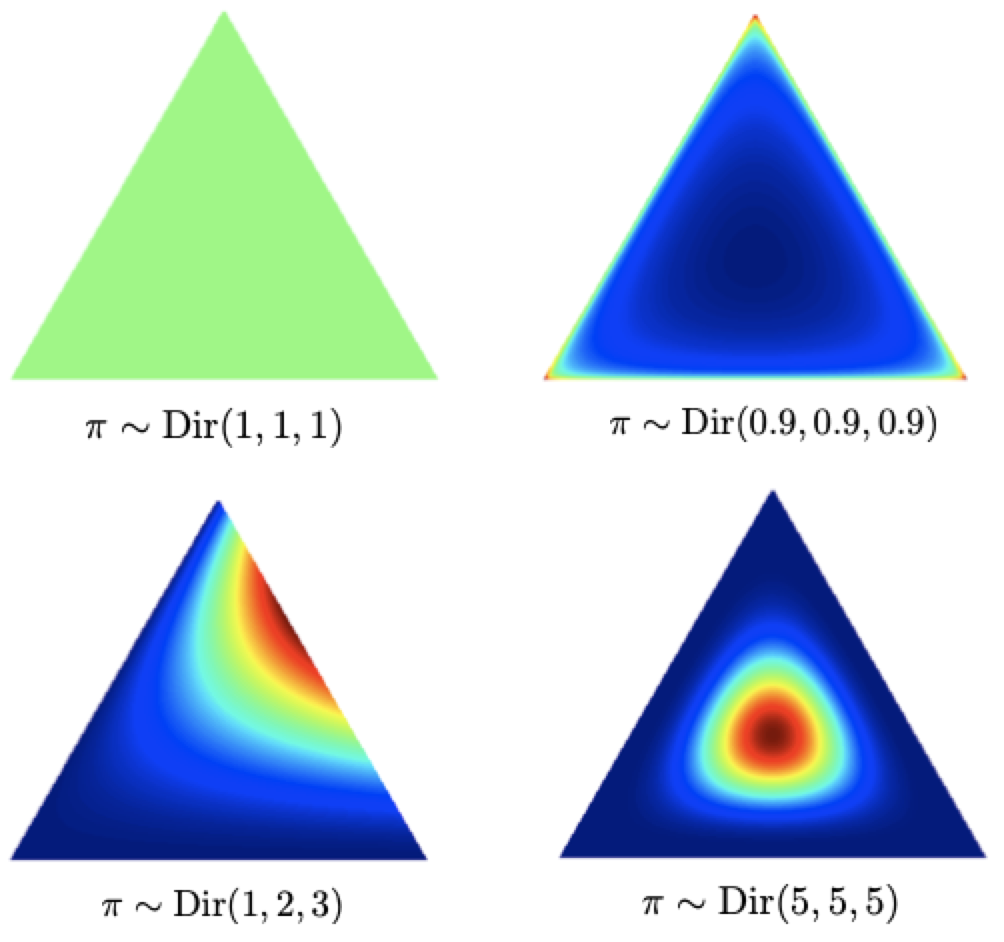}
\caption{Dirichlet distribution as a uniform prior, prior favoring sparse multinomial distribution, a biased prior, and an unbiased prior.}
\label{fig:Dirichlet_dist}
\end{figure}
Figure \ref{fig:Dirichlet_dist} illustrates the Dirichlet distribution with different values of $\alpha$ and $K = 3$ on the simplex $\Pi_2 = (\pi_1, \pi_2, 1- \pi_1-\pi_2)$, showcasing its versatility in representing various probabilistic scenarios.

\subsection{Beta Distribution}

The Beta distribution is a cornerstone of probability theory, providing a continuous probability distribution defined on the interval $[0,1]$. It is characterized by two positive parameters, $a$ and $b$. Interestingly, the Beta distribution serves as a special case of the more general Dirichlet distribution.

Let $\bX$ be a random variable following a Beta distribution, denoted as $\text{Beta}(a, b)$. Its probability density function is given by:

\begin{equation}
\mathbb{P}(d\bx|a, b) = \frac{1}{\beta(a, b)}\bx^{a-1}(1-\bx)^{b-1}d\bx
\end{equation}

where $\beta(a, b) = \frac{\Gamma(a)\Gamma(b)}{\Gamma(a+b)}$, with $\Gamma(\cdot)$ being the gamma function. The parameters $a$ and $b$ govern the shape of the distribution, determining its skewness and kurtosis. 
Properties of the Beta distribution provide insights into its behavior:

\begin{itemize}
    \item The expected value of $\bX$ is $\mathbb{E}(\bX) = \frac{a}{a+b}$, reflecting the balance between the parameters.
    \item The variance of $\bX$ is $Var(\bX) = \frac{a b}{(a+b)^2 (a+b+1)}$, indicating its spread around the mean.
    \item The expected logarithm of $\bX$ is $\mathbb{E}[\ln X] = \frac{\partial \ln \Gamma(a)}{\partial a} - \frac{\partial \ln \Gamma(a+b)}{\partial a} = \psi(a) - \psi (a+b)$, where $\psi$ denotes the digamma function.
\end{itemize}

Moreover, the Beta distribution serves as the conjugate prior for several important distributions including Bernoulli, binomial, negative binomial, and geometric distributions. Its moment generating function further characterizes its statistical properties.

\subsection{Gamma Distribution}
\label{sec:gamma}

The Gamma distribution, a two-parameter family of continuous probability distributions, denoted as $\bX \sim \Gamma(\alpha, \beta)$, extends the Beta distribution to positive real numbers. Here, $\alpha$ and $\beta$ are both greater than zero, shaping the density function as follows:

\begin{equation}
\mathbb{P}(d\bx|\alpha, \beta) = \frac{\beta^\alpha}{\Gamma(\alpha)}\bx^{\alpha-1}\exp{(-\beta\bx)} \mathbbm{1}_{\bx}[0,\infty]d\bx.
\end{equation}

Similar to the Beta distribution, $\Gamma(\cdot)$ represents the gamma function. Notably, the Gamma distribution belongs to the exponential family with natural parameters $\bth = [\alpha-1, -\beta]^T$.

Key properties of the Gamma distribution include:

\begin{itemize}
    \item The expected value of $\bX$ is $\mathbb{E}[\bX] = \frac{\alpha}{\beta}$, indicating its location on the positive real line.
    \item The variance of $\bX$ is $\text{Var}[\bX] = \frac{\alpha}{\beta^2}$, providing a measure of its dispersion.
\end{itemize}

Additionally, the sum of Gamma-distributed random variables follows a Gamma distribution. Specifically, if $\bX_j \sim \Gamma (\alpha_j,\beta)$ for $j=1,2,\dots,N$, then $\sum_{j=1}^{N} \bX_j \sim \Gamma(\sum_{j=1}^{N}\alpha_j,\beta)$.

A particularly interesting property involves the reciprocal of a Gamma-distributed variable. If $\bX \sim \Gamma (\alpha,\beta)$, then $\bY = \frac{1}{\bX}$ follows an inverse-Gamma distribution, denoted as $\text{IG}(\alpha,\beta)$, with its own density function and moments.

\subsection{Student's t-Distribution: A Versatile Tool in Statistical Modeling}

The Student's t-distribution, a cornerstone of statistical theory and practice, finds its origins through two distinct derivations: one as a conjugate prior for the variance of a Gaussian distribution, and the other as the square root of a Gamma random variable \cite{casella2002statistical}. 

Consider a Gaussian random variable $\bX$ with mean $\mu$ and known fixed value, and precision $\tau = 1/\sigma^2$. Placing a prior distribution of $\Gamma(\alpha,\beta)$ over the precision $\tau$ yields a marginal density that follows the Student's t-distribution:

\begin{equation}
\mathbb{P}(d\bx|\mu,\alpha, \beta) = \frac{\Gamma(\alpha+1/2)}{\Gamma(\alpha)(2\pi\beta)^{1/2}}\frac{1}{\left(1+\frac{1}{2\beta}(\bx-\mu)^2\right)^{\alpha+1/2}}d\bx
\end{equation}

This distribution is characterized by the parameters $\mu$, $\alpha$, and $\beta$, with $\Gamma(\cdot)$ representing the gamma function. Notably, when $\alpha = 2$, the Student's t-distribution reduces to the Cauchy distribution, and as $\alpha \to \infty$, it converges to the Gaussian distribution. Often, it's convenient to rewrite the density in terms of $\nu = \alpha/2$ and $\lambda = \alpha/\beta$ for better interpretation and computational efficiency.

On the other hand, the Normal-Inverse-Wishart Distribution is a powerful Bayesian tool for modeling multivariate Gaussian distributions \cite{casella2002statistical, robert2007bayesian}. For a d-dimensional random variable $\bX$ with mean $\mu$ and covariance matrix $\Sigma$, its Gaussian distribution is given by:

\begin{equation}
\mathbb{P}(d\bx|\mu, \Sigma) = \frac{1}{(2\pi)^{d/2}|\Sigma|^{1/2}}\exp\left\{-\frac{1}{2}(\bx-\mu)^T\Sigma^{-1}(\bx-\mu)\right\}d\bx
\end{equation}

This distribution, denoted as $\mathcal{N}(\mu, \Sigma)$, serves as the foundation for many statistical models. Maximum likelihood estimation for the parameters, given observations $\{\bx^n\}_{n=1}^N$, results in estimates for $\mu$ and $\Sigma$.

To facilitate Bayesian inference, conjugate priors are desirable. For the covariance matrix $\Sigma$, the inverse-Wishart distribution, denoted as $\mathcal{IW}(\nu, \boldsymbol{\Psi})$, serves as a conjugate prior. The inverse-Wishart distribution is a multivariate extension of the inverse-Gamma distribution. Its mean and mode are given by:

\begin{equation}
\begin{split}
\mathbb{E}[\Sigma] &= \frac{\boldsymbol{\Psi}}{\nu - d - 1} \quad \text{for} \quad \nu > d + 1\\
\text{mode}(\Sigma) &= \frac{\boldsymbol{\Psi}}{\nu + d + 1}
\end{split}
\end{equation}

For situations where both the mean and covariance matrix are unknown, the Normal-inverse-Wishart distribution, denoted as $\mathcal{NIW}(\mu_0, \lambda, \nu, \boldsymbol{\Psi})$, provides a convenient conjugate prior. This joint distribution combines a Gaussian prior for the mean and an inverse-Wishart prior for the covariance matrix. It is expressed as:

\begin{equation}
\mathbb{P}(d\mu, d\Sigma|\mu_0, \lambda, \nu, \boldsymbol{\Psi}) = \mathcal{N}(\mu; \mu_0, \Sigma/\lambda) \times \mathcal{IW}(\Sigma; \nu, \boldsymbol{\Psi}) d\mu d\Sigma
\end{equation}

This distribution facilitates Bayesian inference by allowing joint estimation of both the mean and covariance matrix from data.

\subsubsection{ Unveiling the Posterior Distribution}

In the realm of Bayesian inference, the posterior distribution stands as a cornerstone, encapsulating the updated beliefs about model parameters after considering observed data. Consider a scenario where $N$ observations $\{\bx^n\}_{n=1}^{N}$ are drawn from a Gaussian distribution $\mathcal{N}(\mu,\Sigma)$. Assume a prior distribution of normal-inverse-Wishart form $\mathcal{NIW}(\mu_0, \lambda, \nu, \boldsymbol{\Psi})$ over parameters $\mu$ and $\Sigma$. Intriguingly, the posterior distribution also adopts a normal-inverse-Wishart structure, albeit with updated hyperparameters $\hat{\mu}$, $\hat{\lambda}$, $\hat{\nu}$, and $\hat{\boldsymbol{\Psi}}$ \cite{hoff2009first, gelman2013bayesian}.

The computation of these hyperparameters involves intricate interplays between prior knowledge and observed data:

\begin{equation}
\begin{split}
\hat{\mu} &= \frac{\lambda\mu_0 + \sum\limits_{n=1}^N \bx^n}{\lambda+N}\\
\hat{\lambda} &= \lambda + N\\
\hat{\nu} &= \nu + N\\
\hat{\boldsymbol{\Psi}} &= \boldsymbol{\Psi} + \mathbb{S} + \frac{\lambda N}{\lambda + N} (\bar{\bx}-\mu_0)(\bar{\bx}-\mu_0)^T
\end{split}
\end{equation}

Here, $\bar{\bx} = \sum_n \bx^n$ denotes the sample mean, and $\mathbb{S} = \sum_n (\bx^n-\bar{\bx})(\bx^n-\bar{\bx})^T$ represents the sample covariance matrix. These updates reflect a synthesis of prior beliefs and newfound evidence, steering the posterior distribution towards a more informed representation of the underlying model.

\subsubsection{Unraveling the Predictive Distribution}

Moving beyond parameter estimation, Bayesian inference facilitates the prediction of future observations through the predictive distribution. Marginalizing over the parameters of the normal-inverse-Wishart distribution, the predictive distribution for a new observation $\bx_{\text{new}}$ assumes a multivariate Student's t-distribution with $(\bar{\nu}-d+1)$ degrees of freedom \cite{hoff2009first, gelman2013bayesian}.

In instances where the normal-inverse-Wishart distribution is proper ($\bar{\nu} > d + 1$), ensuring finite covariance, the predictive density can be approximated for practical computation:

\begin{equation}
p(\bx_{\text{new}}| \{\bx^n\}_{n=1}^{N}, \mu_0, \lambda, \nu, \boldsymbol{\Psi}) \approx \mathcal{N}(\bx_{\text{new}}; \hat{\nu}, \frac{(\hat{\lambda}+1)\hat{\nu}}{\hat{\lambda}(\hat{\nu}-d-1)} \hat{\boldsymbol{\Psi}})
\end{equation}

This approximation, known as the moment-matched Gaussian approximation, furnishes a computationally tractable representation of the posterior distribution, facilitating predictive analysis and decision-making in Bayesian frameworks \cite{gelman2013bayesian, sudderth2006graphical}.

\section{Introduction to Bayesian Nonparametrics}
\label{intro}

In the realm of traditional Bayesian statistics, the framework revolves around the elegant application of Bayes' formula. Given observed data $\mathbf{X}$ with likelihood $\mathcal{L}(\theta)$ and a prior belief encapsulated by $\pi(\theta)$ over the parameters, Bayesian inference calculates the posterior distribution. This foundational approach involves modeling via a prior $\pi(\theta)$ and a likelihood $\mathcal{L}(\theta)$, assuming that data is generated through the following process:

\begin{flalign}
\label{model}
\begin{split}
\theta &\sim \pi(\theta)\\
\mathbf{X}_i & \sim \mathcal{L}(\theta) \quad i = 1, \dots, n
\end{split}
\end{flalign}

This model implicitly acknowledges the conditional independent and identically distributed (i.i.d.) nature of the data, where each observation $\mathbf{X}_i$ is independent of others, conditioned on the underlying parameter $\theta$. The posterior distribution is then derived through Bayes' formula:

\begin{equation}
\label{bayes}
\pi(\theta|\mathbf{X}) = \frac{\pi(\theta)\mathcal{L}(\theta)}{\int \pi(\theta')\mathcal{L}(\theta')d\theta'}
\end{equation}

However, the elegance of Equation \ref{bayes} relies on well-defined densities with respect to a suitable measure, presupposing a finite-dimensional parameter space. Yet, in practice, parameters often retain uncertainty even with a finite number of observations, with Bayesian statistics utilizing the posterior distribution to quantify this uncertainty. This framework encounters limitations when dealing with infinite-dimensional parameter spaces, a gap addressed by Bayesian nonparametric models.

The appeal of Bayesian nonparametrics stems not only from its capacity to learn with increasing data but also from a compelling statistical rationale. This rationale finds its roots in de Finetti's theorem, which concerns infinitely exchangeable sequences of data. A sequence of random variables is deemed infinitely exchangeable if its distribution remains invariant under any finite sequence and permutation. Formally:

\begin{equation}
P(\mathbf{X}_1 \in A_1, \dots, \mathbf{X}_n \in A_n) = P(\mathbf{X}_{\sigma(1)} \in A_1, \dots, \mathbf{X}_{\sigma(n)} \in A_n)
\end{equation}

\begin{theorem}[de Finetti's Theorem]
A sequence $\mathbf{X}_1, \mathbf{X}_2, \dots$ is infinitely exchangeable if and only if, for all $n$, there exists a distribution $G$ such that:
\begin{equation}
P(\mathbf{X}_1 \in A_1, \dots, \mathbf{X}_n \in A_n) = \int_{\Theta} \prod_{j=1}^{n} P(\mathbf{X}_j \in A_j|\theta) G(d\theta)
\end{equation}
\end{theorem}

This theorem asserts the existence of a measure $G$ from which parameters are drawn, ensuring the conditional independence of data. Bayesian nonparametrics thus grapple with fundamental questions:

\begin{enumerate}
\item How do we construct a prior on an infinite-dimensional set?
\item How do we compute the posterior? How do we draw random samples from it?
\item What are the properties of the posterior? Is it consistent? What is its rate of convergence?
\end{enumerate}

In the subsequent sections, we delve into a probability distribution over partitions known as the Chinese restaurant process, demonstrating how its exchangeability property leads to the Dirichlet process. It's worth noting that Bayesian inference methodologies may not align with frequentist approaches, and Bayesian models may lack properties like consistency or optimal convergence rates.

\section{Dirichlet Process}
\label{dp}

For Bayesian nonparametric inference, we venture into placing a prior $\pi$ on infinite-dimensional space. One of the most popular Bayesian nonparametric models over the space of distributions is the Dirichlet process. Ferguson first introduced the Dirichlet process in a seminal paper in 1973 \cite{ferg1973}, leveraging a general theorem by Kolmogorov on the existence of stochastic processes to establish its existence. Despite this definition, the existence of such a process raised measure-theoretic challenges, requiring specific topological conditions on the parameter set, as highlighted by Sethuraman in 1994 \cite{sethuraman1994}. An alternative equivalent definition for the Dirichlet process based on P\'{o}lya urn schemes was introduced by Blackwell and McQueen \cite{blackwell1973ferguson}, while Aldous later presented a process over partitions with the underlying distribution being the Dirichlet process \cite{aldous1985, aldous2006ecole}. Sethuraman further proposed a constructive approach to model the Dirichlet process \cite{sethuraman1994}.

In this discourse, we focus on the following representations: (1) Ferguson's definition of the Dirichlet process \cite{ferg1973}, (2) Stick-breaking process (Sethuraman 1994) \cite{sethuraman1994}, (3) Chinese restaurant process \cite{aldous1985}, (4) Blackwell-MacQueen process (P\'{o}lya urn scheme) \cite{blackwell1973ferguson}.

\subsection{Ferguson Definition of Dirichlet Process}
Ferguson introduced a class of priors that exhibit significant support, enabling the computation of posteriors analytically.

\begin{figure}
\centering
\includegraphics[width = 8cm, height = 3 cm ]{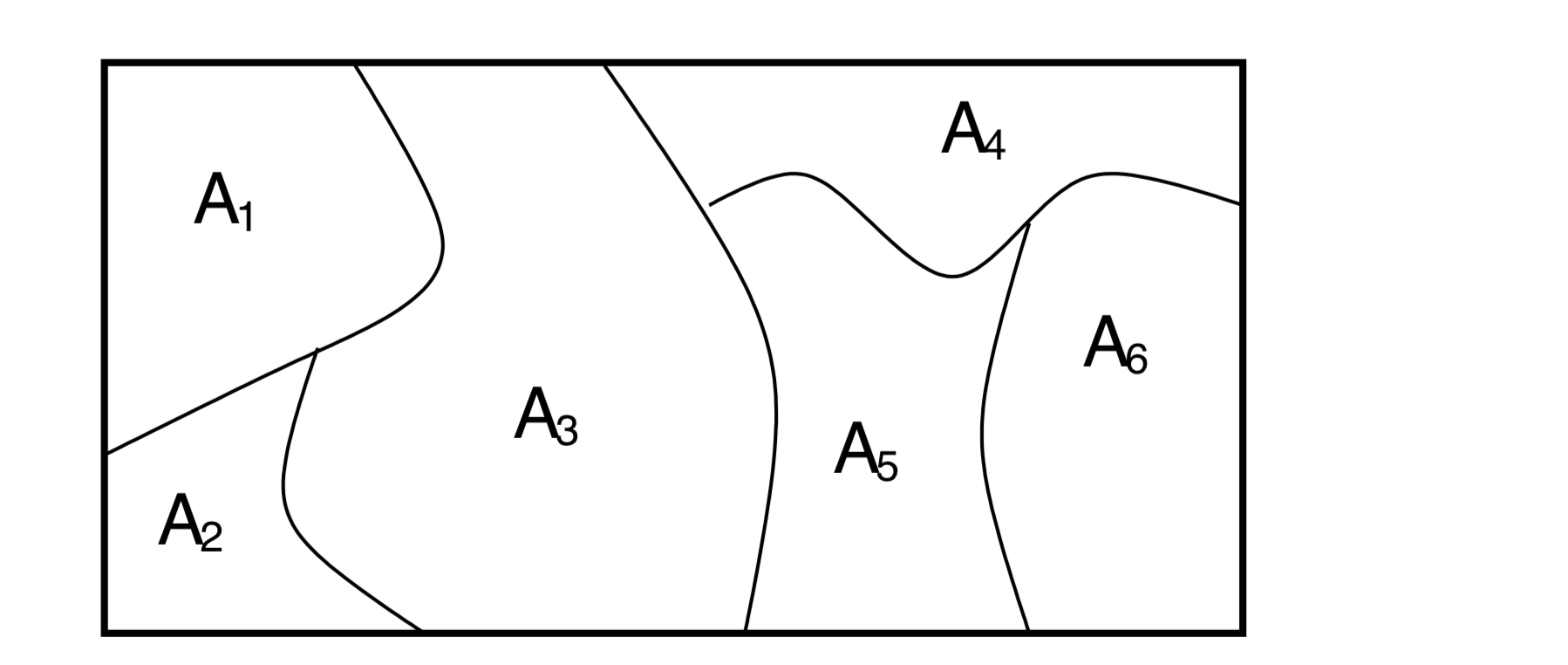}
\caption{Partition of the parameter space.}
\label{fig:partition}
\end{figure}

\textbf{Definition:} The Dirichlet process is a random probability measure over the space $\Theta$ satisfying:
Consider $A_1, \dots, A_n$ as a partition of the Polish space $\Theta$ (refer to Figure \ref{fig:partition}), and let $G \sim DP(\alpha, H)$ be a realization of a Dirichlet process with concentration parameter $\alpha$ and base distribution $H$. Then, (a) $G$ is a random measure;
    (b) $G$ is discrete with probability one;
    (c) The vector $(G(A_1), \dots, G(A_n))$ is a probability vector;
    (d) $(G(A_1), \dots, G(A_n)) \sim \text{Dirichlet}( \alpha H(A_1), \dots, \alpha H(A_n))$.
    
It's evident that $G$ is a random measure; hence $G(A)$ is a random variable given an event $A$. The following holds from the definition:

\begin{flalign}
\begin{split}
\mathbb{E}[G(A)] &= H(A)\\
\text{Var}[G(A)] &= \frac{H(A)(1-H(A))}{\alpha+1}
\end{split}
\end{flalign}

\subsection{Posterior Distribution of Dirichlet Process}
In this section, we delve into obtaining the posterior distribution of a Dirichlet process. Assuming the model in Equation \ref{model} with the Dirichlet process as the prior, Ferguson proves that the posterior distribution is also a Dirichlet process \cite{ferg1973}.

\begin{theorem}
Consider the following hierarchy:
\begin{flalign}
\begin{split}
G &\sim DP(\alpha, H)\\
\theta_j | G &\sim G \quad j = 1, \dots, n
\end{split}
\end{flalign}

Then, the posterior distribution is $DP(\alpha+n, \frac{1}{\alpha+n}\sum\delta_{\theta_i}+ \frac{\alpha}{\alpha+n}H)$ \cite{orbanz2010bayesian, hjort2010bayesian, ghosal1999, barron1999}.
\end{theorem}

\subsection{A Constructive Method: Stick-Breaking Construction}

\begin{figure}
\centering
\includegraphics[width=7cm,height=4cm]{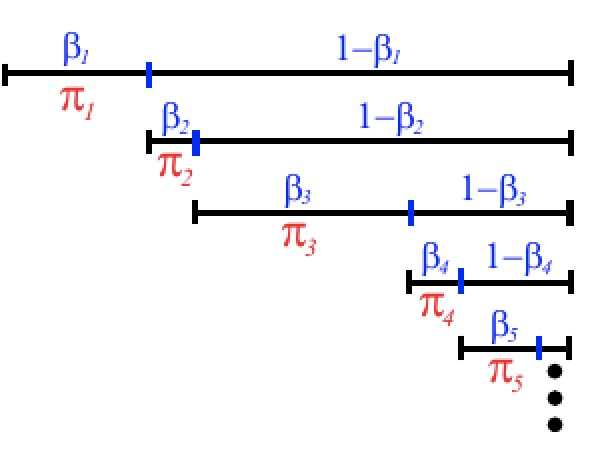}
\caption{Stick-breaking process}
\label{fig:stick}
\end{figure}
 
Ferguson's original formulation of the Dirichlet process offers a theoretical foundation but lacks practicality in direct sampling. Sethuraman introduced a pragmatic method to draw from a Dirichlet process, known as the stick-breaking construction. This method provides a tangible procedure for generating a single random distribution $G$ from a Dirichlet process $\text{DP}(\alpha, H)$, following a hierarchical approach:

\begin{flalign}
\begin{split}
\theta_j &\stackrel{i.i.d.}{\sim} H\\  
\pi_j &\sim \text{GEM}(\alpha), \\
G &= \sum\limits_{j=1}^{\infty} \pi_j \delta_{\theta_j}
\end{split}
\end{flalign}

Here, GEM$(\alpha)$ denotes the Griffiths-Engen-McCloskey distribution, defined as:

\begin{equation}
\begin{split}
\beta_j &\stackrel{i.i.d.}{\sim} \text{Beta}(1,\alpha)\\
\pi_j &= \beta_j \prod\limits_{i=1}^{j-1} (1- \beta_i)
\end{split}
\end{equation}

This process earns its moniker, the stick-breaking process, from the analogy of progressively dividing a unit-length stick to construct the weights $\pi_j$'s. Imagine starting with a stick of unit length. The first weight, $\pi_1$, is derived by breaking the stick at a random point $\beta_1$. The remaining stick now measures $1-\beta_1$. Subsequently, the next weight $\pi_2$ is determined by breaking a proportion $\beta_2$ from the remaining stick, and so forth, generating the entire sequence of weights $\pi_j$'s. Figure \ref{fig:stick} illustrates this iterative procedure.

The stick-breaking construction method for drawing from a Dirichlet process offers practicality and insight into the underlying structure of Bayesian nonparametric models. After sequentially dividing a unit-length stick to form weights $\pi_j$'s, the resulting distribution $G$ is inherently discrete, providing a discrete representation of the Dirichlet process with probability one. This discrete nature simplifies computational procedures and aligns with the discrete nature of many real-world phenomena, making it particularly useful for modeling discrete data or processes with countable outcomes.

Furthermore, the requirement that the weights $\pi_j$'s sum up to one ensures the coherence and validity of the resulting probability distribution. This property maintains the integrity of the overall probability structure, allowing for meaningful probabilistic interpretations and facilitating straightforward applications in Bayesian inference. Consequently, practitioners can rely on the stick-breaking construction to generate random distributions that adhere to the fundamental principles of probability theory, thus enhancing the reliability and utility of Bayesian nonparametric methods in diverse applications.

Despite the decreasing trend in the average values of the weights $\pi_j$'s, their individual ordering remains non-strictly decreasing. This characteristic introduces complexity in modeling and analysis, prompting the exploration of alternative approaches. One such approach, the Poisson-Dirichlet process introduced by Kingman \cite{kingman1967completely}, aims to order the weights but often leads to computational intractability due to its intricacies. Thus, while acknowledging the non-strictly decreasing nature of the weights, practitioners must carefully balance the desire for orderliness with computational feasibility when employing Bayesian nonparametric models.

\begin{figure}
\centering
\includegraphics[width = 10cm, height = 5cm ]{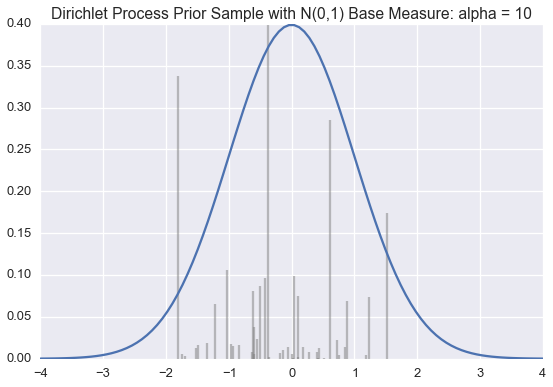} 
\caption{ A draw from a Dirichlet process with Gaussian mean}
\label{fig:dp}
\end{figure}

Figure \ref{fig:dp} provides a visual depiction of a draw from a Dirichlet process with Gaussian mean, showcasing the practical application of the stick-breaking construction. This method not only facilitates sampling from Dirichlet processes but also offers insights into the underlying structure of the process, aiding in the comprehension and utilization of Bayesian nonparametric models.

\subsection{Dirichlet Process Mixture Model}
The Dirichlet process, while not suitable as a prior for directly estimating densities due to its generation of a discrete random measure, can be extended for density estimation through the Dirichlet process mixture model. In this model, we aim to estimate the density \( f \) of a distribution \( F \) by employing a Dirichlet process. Suppose we have independent and identically distributed (i.i.d.) data \( \mathbf{X}_1, \mathbf{X}_2, \dots \) drawn from a distribution \( F \) with density \( f \). To estimate \( f \), we place a Dirichlet process on the parameter space and draw parameters from the mean of this Dirichlet process. Each parameter is selected with some probability according to the Griffiths-Engen-McCloskey (GEM) distribution with parameter \( \alpha \), forming an infinite mixture model known as the Dirichlet process mixture model. This model mirrors the random distribution \( F \) generated by the Dirichlet process but smoothens out the point mass distributions \( \delta_{\theta_j} \) with densities \( f(\cdot|\theta_j) \), \cite{antoniak1974, ghosal2007}.

The construction of the Dirichlet process mixture model follows a hierarchical approach. We first draw a random distribution \( G \) from the Dirichlet process \( \text{DP}(\alpha, H) \). Then, we draw parameters \( \theta_j \) independently and identically distributed from \( G \), and subsequently draw observations \( \mathbf{X}_j \) from the corresponding densities \( f(\cdot|\theta_j) \).  To simplify the model and focus on the mean and concentration parameters, we can marginalize out \( G \), resulting in a hierarchical model where parameters are drawn directly from the base distribution \( H \) and assigned to clusters according to a categorical distribution. Bayesian inference methods such as Markov chain Monte Carlo (MCMC) \cite{escobar1994estimating, ishwaran2000markov, jain2004split,maceachern1994estimating, williamson2013parallel}or variational Bayes methods are commonly employed for inference in this model \cite{ kurihara2007collapsed, wang2011online, blei2006variational}. To simplify the model and make it depend only on the mean and concentration parameters, we marginalize out \( G \). This results in the following hierarchical structure:

\begin{align}
    \begin{split}
        \pi |\alpha & \sim \text{GEM}(\alpha) \\
        \theta_j | H & \stackrel{i.i.d.}{\sim} H \\
        z_j|\pi & \sim \text{Cat}(\pi) \\
        X_j | \Theta, z_j & \sim f(\cdot | \theta_{z_j})
    \end{split}
\end{align}

Here, \( \text{Cat}(\pi) \) denotes the categorical distribution with parameter \( \pi \). Bayesian inference techniques such as Markov Chain Monte Carlo (MCMC) or variational Bayes methods are commonly employed for inference \cite{neal2000, escobar1995}.

\subsection{Dirichlet Process and Clustering: Chinese Restaurant Process}
\label{sec:crp}
The Chinese Restaurant Process (CRP) provides a foundational framework for understanding the behavior of the Dirichlet process mixture model. For a fixed \( \alpha > 0 \), the CRP(\(\alpha\)) is a distribution over all partitions of a labeled set \( [n] := \{1,2,...,n\} \), where each subset of the partition represents a table or cluster. In this metaphorical restaurant setting, each customer (data point) enters and selects a table according to a set of probabilities based on the existing table occupancy. The CRP exhibits exchangeability, and the induced distribution over partitions, known as the exchangeable partition probability function (EPPF), provides insight into the distribution of cluster sizes. While the CRP is not exchangeable in the de Finetti sense, it establishes a close relationship between partition exchangeability and sequence exchangeability, paving the way for the construction of random sequences from random partitions. This construction ultimately leads to the realization of a de Finetti exchangeable sequence, implying the existence of a random probability measure under which the data is distributed i.i.d., with the underlying distribution being a Dirichlet process, as expounded by Aldous in 1985 \cite{aldous2006ecole}.

\begin{figure}[t]
\centering
\includegraphics[width=8.5cm, height=2cm]{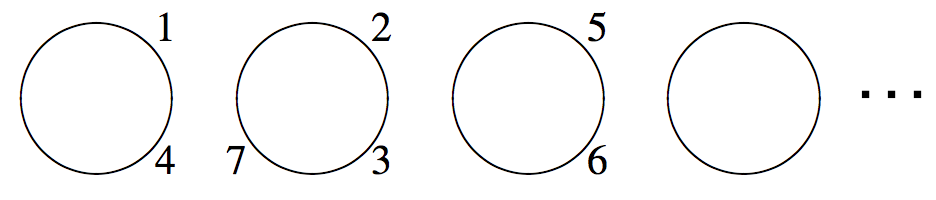}
\caption{Chinese restaurant process.}
\label{fig:crp}
\end{figure}

The Chinese Restaurant Process (CRP), denoted as CRP($\alpha$), is a distribution over all partitions of the labeled set $[n]$, where $\rho \sim \text{CRP}(\alpha)$ represents a partition over $[n]$. In this process, each subset of the partition can be conceptualized as a table or a cluster, with the data often referred to as customers. The distribution is recursively defined, where each customer, as illustrated in Figure \ref{fig:crp}, enters the restaurant and randomly selects a table.

The probability of choosing a particular table $\mathcal{C}$ is given by:
\begin{equation}
\begin{split}
\mathbb{P}(\text{Choose table}\hspace{0.1cm} \mathcal{C}) &= \frac{n_{\mathcal{C}}}{\alpha + \sum_{\rho} n_{\mathcal{C}}} \\
\mathbb{P}(\text{Choose a new table}) &= \frac{\alpha}{\alpha + \sum_{\rho} n_{\mathcal{C}}}
\end{split}
\end{equation}

Here, $n_{\mathcal{C}}$ represents the number of customers at table $\mathcal{C}$. The CRP is exchangeable, and the induced distribution over partitions (without labeling) is referred to as the exchangeable partition probability function (EPPF):

\begin{equation}
\mathbb{P}(n_1, \dots, n_K|\alpha) = \frac{\alpha^{K}}{\alpha^{[n]}} \prod_j (n_j - 1)!
\end{equation}

where $\alpha^{[n]} = \alpha (1+ \alpha) \dots (\alpha + n -1)$. Although this process is not exchangeable in the de Finetti sense, there exists a close relationship between partition exchangeability and sequence exchangeability.

To elaborate, for each $\mathcal{C} \in \rho$, we define $\bth^\star_{\mathcal{C}} \sim H$, and for each $j \in [n]$, we define $\bth_j = \bth^\star_{\mathcal{C}}$, where $\mathcal{C} \in \rho$ and $j \in \mathcal{C}$. Thus, $\bth_1, \bth_2, \dots$ are de Finetti exchangeable. 

In summary, the construction of a random sequence from a random partition follows the hierarchy:
\begin{align}
\begin{split}
    \rho &\sim \text{CRP} (\alpha) \\
    \bth^\star_{\mathcal{C}} &\sim H \quad \text{for each} \quad \mathcal{C}\in \rho \\
    \bth^\star_{j} &= \bth^\star_{\mathcal{C}} \quad \text{for each} \quad j \in [n], \quad \mathcal{C} \in \rho, \quad j \in  \mathcal{C}
\end{split}
\end{align}

\begin{theorem}(Aldous 1985)
The sequence $\bth_1, \bth_2, \dots$ generated as described above is de Finetti exchangeable. Consequently, there exists a random probability measure under which the data is distributed i.i.d., with the underlying distribution being the Dirichlet process.
\end{theorem}

\subsection{The Blackwell-MacQueen Distribution: P\'{o}lya Urn Scheme}

The Blackwell-MacQueen distribution serves as a generalization of the P\'{o}lya urn, offering a framework that encapsulates the Chinese Restaurant Process (CRP) model introduced in Section \ref{sec:crp}. Let $\bth^*_1, \dots, \bth^*_K$ denote the unique cluster parameters. In this context, the predictive distribution is given by:

\begin{equation}
\bth_{n+1} | \bth_{1}, \dots, \bth_{n} \sim \frac{1}{\alpha+n}\sum \delta_{\bth_j} + \frac{\alpha}{\alpha+n}H
\end{equation}

Here, $\delta_{\bth_j}$ represents the point mass at $\bth_j$. The distribution over the sequence of $\bth$ follows the Blackwell-MacQueen distribution. This formulation elegantly captures the evolution of cluster parameters as new data points are incorporated into the model. The parameter $\alpha$ plays a crucial role in balancing the influence between the observed data and the prior knowledge encapsulated by $H$.

\subsubsection{Applications of Dirichlet Process}
The Dirichlet Process Mixture Model serves as a versatile tool across a wide range of disciplines, including tracking, bioinformatics, NLP, image analysis, and beyond owing to its flexibility and ability to handle unknown or varying numbers of underlying components within data. In tracking applications, DPMMs are utilized to model the trajectories of moving objects, such as vehicles or pedestrians, in dynamic environments. By flexibly adapting the number of clusters to the changing number of objects and their trajectories over time, DPMMs provide robust and accurate tracking results \cite{caron2007bayesian, fox2011, caron2012}.

In bioinformatics, DPMMs play a crucial role in analyzing biological data, such as DNA sequences, gene expression profiles, and protein structures. By clustering sequences or expression profiles into groups with similar characteristics, DPMMs aid in identifying genetic patterns, protein interactions, and disease biomarkers, facilitating advancements in personalized medicine and drug discovery \cite{nguyen2013dirichlet, rasmussen2008modeling,brown2008efficient,caron2007bayesian}.

In natural language processing (NLP), DPMMs are employed for tasks such as document clustering, topic modeling, and sentiment analysis. By clustering documents or words into latent topics, DPMMs enable the discovery of underlying themes and sentiments within large text corpora, enhancing information retrieval, document summarization, and sentiment classification \cite{dreyer2011discovering, chen2015hidden}.

In image analysis and computer vision, DPMMs find applications in image segmentation, object recognition, and scene understanding. By grouping pixels or image patches into coherent regions or objects, DPMMs facilitate the extraction of meaningful structures and features from images, enabling tasks such as object tracking, scene reconstruction, and content-based image retrieval \cite{yakhnenko2008annotating,ding2011interactive, ma2018variational}.

\subsection{Hierarchical Dirichlet Process Modeling}

In statistics, a common objective is to group data while preserving dependencies among these groups, allowing them to share statistical strength \cite{Teh2006}. Traditional hierarchical modeling in Bayesian statistics achieves this by leveraging hierarchical structures to maintain group dependencies, with shrinkage effects manifested through posterior distributions. The Hierarchical Dirichlet Process (HDP) represents one such hierarchical model, forming a hierarchy of Dirichlet processes.

However, in hierarchical Bayesian nonparametrics, the challenge lies in placing a prior on a prior, particularly ensuring the discreteness of the prior. To address this, the discreteness property of the Dirichlet process is exploited by placing a Dirichlet process as the prior over the prior. By introducing an additional level of Dirichlet process, the HDP shares countably infinite cluster identities among groups, enabling the representation of unique cluster propositions. Consequently, HDP mixture models accurately capture the characteristics of grouped data.

The Hierarchical Dirichlet Process (HDP) mixture model offers a powerful framework for clustering data into multiple groups while maintaining dependencies among these groups. In this model, the base distribution \( H \) serves as the foundation, influencing both the global distribution \( G_0 \) and the local distributions \( G_1 \) and \( G_2 \) associated with each group. Within each group, the parameters \( \theta_{j,m} \) capture the characteristics of individual observations, sampled from their respective local distributions. This hierarchical structure facilitates the sharing of statistical strength across groups while accommodating variability within each group. Consequently, the HDP mixture model provides a flexible and scalable approach for analyzing complex datasets, allowing for nuanced insights into the underlying patterns and relationships among the observations. The HDP is mathematically formulated as follows:
\begin{flalign}
\label{eq:hdp}
\begin{split}
& G_0 \sim \text{DP}(\gamma, H)\\
& G_m | G_0 \sim \text{DP}(\alpha, G_0) \\
& \bth_{j,m} | G_m \sim G_m
\end{split}
\end{flalign}

It's worth noting that there are several statistically equivalent constructions of the HDP. One constructive approach is based on the generalization of the Chinese restaurant process, known as the Chinese restaurant franchise \cite{Teh2006}. Moreover, the hierarchical structure in Equation \ref{eq:hdp} allows for density estimation using an infinite mixture model:
\begin{equation}
\bX_{j,m} | \bth_{j,m} \sim f(\cdot | \bth_{j,m}).
\end{equation}

This formulation of the HDP facilitates Bayesian inference using methods such as Markov chain Monte Carlo (MCMC) or variational Bayes, enabling the estimation of model parameters and latent variables from observed data \cite{neal2000,escobar1995}. 

The application of hierarchical Dirichlet processes (HDP) spans a wide array of domains, with tracking being a particularly notable area of interest. In tracking applications, HDP models are leveraged to handle complex data structures inherent in tracking scenarios, such as the varying number of targets, their trajectories, and associated uncertainties. By employing HDP models, tracking systems can effectively handle the dynamic nature of the tracking environment, adapt to changes in the target behavior, and provide accurate predictions in real time. Beyond tracking, HDP finds applications in natural language processing, topic modeling, image segmentation, and clustering tasks. In these contexts, HDP offers a flexible and scalable framework for capturing latent structures and patterns in data, making it a versatile tool for various machine learning and statistical applications \cite{dai2014supervised, li2016hierarchical, manouchehri2021batch, calderon2015inferring, moraffah2019use, fox2007hierarchical, gao2011tracking, liu2024shared}.

\subsection{Chinese Restaurant Franchise}

The Chinese Restaurant Franchise (CRF) construction is a constructive approach to understanding the Hierarchical Dirichlet Process (HDP) and its hierarchical nature \cite{Teh2006, teh2005sharing}. It provides a framework for generating distributions from an HDP by recursively defining distributions at each level of the hierarchy.

Let $H_0$  denote the base distribution at the global level and the base distribution at the local level as $H_1$. The CRF construction defines the generative process for sampling from an HDP as follows:
\begin{enumerate}
\item At the global level:
   Assume a CRP($\gamma$) process for customer seating arrangements at the global restaurant. Each table corresponds to a local restaurant.

\item At the local level:
   For each table (local restaurant) sampled from the global CRP, sample a distribution $G_m$ from a DP($\alpha, H_0$). This distribution represents the menu at the local restaurant.
    Customers at each local restaurant follow a CRP($\alpha$) process to choose tables (sub-local restaurants) within the local restaurant. Each sub-local restaurant corresponds to a table in the global restaurant.

\item  At the sub-local level:
    For each table (sub-local restaurant) sampled from the local CRP, sample a distribution $G_{m,k}$ from a DP($\beta, G_m$). This distribution represents the menu at the sub-local restaurant.
    Customers at each sub-local restaurant follow a CRP($\beta$) process to choose dishes within the sub-local restaurant. Each dish corresponds to a table in the local restaurant.
\end{enumerate}
This process continues recursively, defining distributions at each level of hierarchy. The resulting hierarchy of Dirichlet processes allows for the representation of nested dependencies among groups, providing a flexible framework for hierarchical Bayesian nonparametric modeling.
The CRF construction involves defining the probability distributions and sampling procedures at each level of the hierarchy, as described above. These distributions are based on the Chinese Restaurant Process (CRP) and the Dirichlet Process (DP), incorporating concentration parameters $\gamma$, $\alpha$, and $\beta$, as well as base distributions $H_0$ and $H_1$. Sampling from these distributions involves probabilistic procedures such as CRP sampling and DP sampling, which can be implemented using appropriate algorithms such as Gibbs sampling or Metropolis-Hastings sampling.

\subsection{Nested Dirichlet Process}

Nested Dirichlet Process (nDP) is an extension of the Hierarchical Dirichlet Process (HDP) that allows for more flexible modeling of hierarchical structures in Bayesian nonparametrics \cite{rodriguez2008nested}. It introduces an additional layer of hierarchy, enabling the representation of nested dependencies among groups.

Similar to HDP, nDP involves a hierarchy of Dirichlet processes, but with an added level of nesting. Let's denote $G_0$ as the global distribution, $G_1, G_2, \ldots$ as local distributions for each group, and $G_{1,1}, G_{1,2}, \ldots$ as sub-local distributions within each local group. The model can be formulated as follows:

\begin{flalign}
\begin{split}
& G_0 \sim \text{DP}(\gamma, H_0)\\
& G_m | G_0 \sim \text{DP}(\alpha, G_0) \\
& G_{m,k} | G_m \sim \text{DP}(\beta, G_m) \\
& \bth_{j,m,k} | G_{m,k} \sim G_{m,k}
\end{split}
\end{flalign}

Here, $H_0$ represents the base distribution at the global level, $\gamma$ controls the concentration parameter for the global distribution, $\alpha$ controls the concentration parameter for local distributions, and $\beta$ controls the concentration parameter for sub-local distributions. $\bth_{j,m,k}$ represents the parameters associated with each observation in the $k$-th sub-group of the $m$-th local group.

Inference in nested Dirichlet process models can be challenging due to the complexity introduced by multiple levels of hierarchy. Markov chain Monte Carlo (MCMC) methods, such as Gibbs sampling or Metropolis-Hastings algorithms, are commonly employed for posterior inference. These methods iteratively sample from the posterior distribution of model parameters and latent variables given the observed data \cite{rodriguez2008nested}.

Variational inference provides an alternative approach for approximate posterior inference in nDP models \cite{wang2009variational}. By approximating the posterior distribution with a simpler distribution from a tractable family, variational inference aims to minimize the Kullback-Leibler divergence between the true posterior and the approximating distribution.

\subsubsection{Applications}

Nested Dirichlet Process models find applications in various domains such as natural language processing, topic modeling, and clustering. Their ability to capture hierarchical structures makes them suitable for modeling complex data with nested dependencies. By incorporating additional layers of hierarchy, nDP models offer enhanced flexibility and expressiveness in capturing the underlying structure of data.

\subsection{Hierarchical Nested Dirichlet Process (HNDP)}

The Hierarchical Nested Dirichlet Process (HNDP) extends the Hierarchical Dirichlet Process (HDP) by introducing nested levels of clustering, enabling the modeling of data with multiple levels of hierarchy \cite{paisley2014nested}. This hierarchical structure allows for the creation of clusters within clusters, providing a more nuanced representation of complex datasets.

At the core of the HNDP is a recursive formulation that combines multiple levels of Dirichlet processes. Let \( H_0 \) be the base distribution, representing the top-level global distribution. At each level of the hierarchy, a local distribution is defined, denoted by \( G_m \), where \( m \) indexes the level of nesting. The base distribution \( H_0 \) influences the top-level global distribution \( G_0 \), which serves as the starting point for subsequent levels of clustering. It captures the overarching characteristics shared among all clusters. At each level \( m \), the local distribution \( G_m \) is influenced by the preceding level's distribution. This hierarchical dependency allows for the creation of nested clusters, where each cluster at level \( m \) is associated with a specific cluster in the previous level.

For each observation \( j \) within each cluster \( m \), parameters \( \theta_{j,m} \) are sampled from the corresponding local distribution \( G_m \). These parameters capture the unique characteristics of the observations within each cluster, accounting for both intra-cluster similarity and inter-cluster variability. The HNDP model can be recursively defined as follows:

\begin{flalign*}
& G_0 \sim DP(\gamma, H_0) \\
& G_m | G_{m-1} \sim DP(\alpha_m, G_{m-1}) \quad \text{for } m = 1, 2, \ldots \\
& \theta_{j,m} | G_m \sim G_m \quad \text{for } j = 1, 2, \ldots, n_m
\end{flalign*}

where \( DP(\cdot) \) denotes the Dirichlet Process, \( \gamma \) and \( \alpha_m \) are concentration parameters, and \( n_m \) represents the number of observations within each cluster at level \( m \).

Inference in the HNDP model typically involves Bayesian methods such as Markov chain Monte Carlo (MCMC) or variational inference. These methods aim to estimate the posterior distribution of the model parameters, allowing for the identification of meaningful clusters and hierarchical structures within the data.

\subsubsection{Applications}

The application of hierarchical nested Dirichlet processes (HNDP) spans various fields, demonstrating its versatility and effectiveness in modeling complex hierarchical structures within data. In natural language processing (NLP), HNDP models are utilized for tasks such as document clustering, topic modeling, and sentiment analysis. By hierarchically organizing words into topics, subtopics, and sub-subtopics, HNDP models capture the inherent hierarchical nature of language, allowing for more interpretable and contextually rich representations of text data. In bioinformatics, HNDP finds application in analyzing biological sequences, such as DNA and protein sequences, where hierarchical structures exist at multiple levels, including nucleotide/amino acid sequences, genes, and functional domains. By modeling these hierarchical relationships, HNDP enables more accurate prediction of biological properties and interactions. Additionally, in recommendation systems and collaborative filtering, HNDP can capture hierarchical preferences and relationships among users, items, and categories, leading to more personalized and effective recommendations. Overall, the application of hierarchical nested Dirichlet processes extends across various domains, offering a powerful framework for modeling complex hierarchical structures and relationships within data.

\subsection{The Nested Chinese Restaurant Process}

The Nested Chinese Restaurant Process (NCRP) is a probabilistic model used for hierarchical clustering, allowing for the creation of nested structures within clusters \cite{blei2010nested, griffiths2003hierarchical}. It extends the metaphor of customers choosing tables in a Chinese restaurant to multiple levels of nesting, providing a flexible framework for capturing complex data relationships.

The construction of the Nested Chinese Restaurant Process involves a recursive process where customers (representing data points) choose tables (representing clusters) at each level of nesting. At each level, the probability of choosing a new table versus joining an existing one depends on the concentration parameter and the number of customers already seated at each table. At the base level, the top-level Chinese restaurant represents the global distribution of observations. Each customer enters the restaurant and selects a table according to the distribution of existing tables and a concentration parameter $\gamma$. This process generates clusters at the global level, capturing the overall structure of the data.
To introduce nested clustering, each table at the global level becomes a Chinese restaurant at the next level of nesting. Customers entering these restaurants now choose tables within the parent clusters, creating nested sub-clusters. This recursive process continues for each level of nesting, allowing for the generation of clusters within clusters. Consider a dataset with \(n\) observations, the nested Chinese restaurant process is defined as
 \begin{enumerate}
\item Base Level (Level 1):
   At the base level, each customer \(j = 1, 2, \ldots, n\) enters the restaurant.
    The probability that customer \(j\) joins an existing table \(k\) is proportional to the number of customers already seated at table \(k\) and a concentration parameter \(\gamma\), denoted as \(m_k\).
   The probability that customer \(j\) chooses a new table is proportional to \(\gamma\). In other words, each customer \( j \) selects a table \( G_{0,j} \) according to the distribution \( DP(\gamma, G_0) \), where \( G_0 \) is the base distribution.

   The probability distribution for customer \(j\) choosing table \(k\) at the base level is given by:
   \[ P(z_{j,1} = k) = \frac{m_k}{\gamma + j - 1} \]

   where \(z_{j,1}\) represents the table assignment for customer \(j\) at level 1.

\item  Nested Levels (Level \(m > 1\)):
    At each nested level, the tables from the previous level become "restaurants" where customers choose tables according to a similar process as the base level.
   The concentration parameter at each level (\(\gamma_m\)) influences the probability of creating new clusters versus joining existing ones.

   The probability distribution for customer \(j\) choosing table \(k\) at level \(m\) is given by:
   \[ P(z_{j,m} = k | z_{j,m-1} = l) = \begin{cases} 
       \frac{m_k}{\gamma_m + m_l - 1} & \text{if } k = l \\
       \frac{\gamma_m}{\gamma_m + m_l - 1} & \text{if } k \neq l 
   \end{cases} \]

   where \(z_{j,m}\) represents the table assignment for customer \(j\) at level \(m\), and \(m_k\) is the number of customers already seated at table \(k\) at level \(m\). In other words,  each table \( G_{m-1,j} \) at level \( m-1 \) becomes a Chinese restaurant. Customers within each cluster \( G_{m-1,j} \) select tables at level \( m \) according to \( DP(\gamma_m, G_{m-1,j}) \), where \( \gamma_m \) is the concentration parameter for level \( m \).

\end{enumerate}

Inference in the Nested Chinese Restaurant Process typically involves Bayesian methods such as Markov chain Monte Carlo (MCMC) or variational inference. These methods aim to estimate the posterior distribution of the model parameters, including the concentration parameters \(\gamma\) and \(\gamma_m\), allowing for the identification of meaningful nested clusters within the data \cite{wang2009variational}.

\subsubsection{Applications}

The Nested Chinese Restaurant Process has applications in various fields, including natural language processing, image analysis, and social network analysis. It enables the modeling of complex data structures with nested clustering patterns, providing insights into hierarchical relationships among observations.

\subsection{Dependent Dirichlet Process}

The Dirichlet Process is a fundamental construct in Bayesian nonparametric statistics, serving as a prior distribution for random measures. It allows for an infinite-dimensional parameter space, making it particularly useful for modeling situations where the number of clusters or components in the data is unknown and may grow with the sample size. However, in many real-world applications, data exhibit dependencies that cannot be adequately captured by traditional Dirichlet Processes, which assume independence among observations. The Dependent Dirichlet Process (DDP) extends the DP framework to account for such dependencies, making it a powerful tool for modeling complex relational structures within data \cite{Mac2000,mac2000unpublished, Mac99, quintana2022dependent, maceachern2016nonparametric}.

Let us consider a collection of random probability measures indexed by a continuous parameter space, denoted by \(G = \{G(\cdot;\theta), \theta \in \Theta\}\), where \(\Theta\) is the parameter space. The DDP defines a prior distribution over this collection of measures, allowing for dependencies between measures corresponding to different values of \(\theta\). 

Formally, let \(G_0\) be a base measure and \(H\) be a measure over a space \(\mathcal{X}\). A dependent Dirichlet Process \(G\) is defined as:

\[G \sim \text{DP}(c G_0 + (1-c) H)\]

where \(c\) is a dependence parameter in the range \([0,1]\). When \(c = 0\), the DDP reduces to the standard Dirichlet Process, while \(c = 1\) implies complete dependence among the measures.

The hierarchical construction of the DDP involves introducing a latent variable \(Z\), which represents the cluster assignment for each observation. Let \(G_0\) be the base measure of the DP, and for each cluster \(k\), let \(G_k\) be a dependent measure sampled from a base measure \(G_0\) and a dependence measure \(H\). The DDP is then defined by:

\[G = \sum_{k=1}^{\infty} \pi_k \delta_{\theta_k}\]

where \(\pi_k\) are the mixing proportions, \(\delta_{\theta_k}\) are point masses at locations \(\theta_k\), and \(\{\theta_k\}\) are drawn from \(G_0\) or \(H\) depending on the value of \(Z\). The choice of dependence measure \(H\) allows for flexibility in modeling various types of dependencies. For instance, one may use a Gaussian process prior to induce smooth dependencies between measures, or a Markov chain prior to capture sequential dependencies. Inference in dependent Dirichlet processes typically involves Markov chain Monte Carlo (MCMC) methods, such as Gibbs sampling or Metropolis-Hastings, to sample from the posterior distribution of the parameters and latent variables \cite{Mac1998, tierney1994, escobar1995, escobar1994estimating}.

In a general sense, the dependent Dirichlet process (DDP), initially proposed by MacEachern \cite{Mac2000,mac2000unpublished, Mac99}, has paved the way for the development of the DDP mixture model (DDPMM). This model generalizes the Dirichlet Process Mixture Model (DPMM) by incorporating birth, death, and transition processes for the clusters within the model. Moreover, low-variance approximations to DDPMM have been devised, leading to the creation of dynamic clustering algorithms.

In a time-varying context, it becomes intuitive to introduce distinct Dirichlet Process (DP) priors for different time steps. The generative model for such scenarios can be expressed as follows:

\[D_t \sim \text{DP}(\alpha, H_t)\]
\[ \theta_{t,i} \mid D_t \sim D_t \text{ for } i=1, \ldots, n_t, \text{ and } t=0, \ldots, T \]
\[ X_{t:i} \mid \theta_{t,i} \sim F(\theta_{t:i}) \text{ for } i=1, \ldots, n_t, \text{ and } t=0, \ldots, T \]

Here, \(D_t\) represents the DP prior at time \(t\), with concentration parameter \(\alpha\) and base measure \(H_t\). The parameters \(\theta_{t,i}\) are drawn from \(D_t\) for each observation \(i\) at time \(t\), and the data \(X_{t:i}\) are generated from a distribution \(F\) parameterized by \(\theta_{t:i}\).

\subsubsection{Dependent Dirichlet Processes based on Poisson Processes}
A Poisson-based construction of the DDP leverages the connection between Poisson and Dirichlet processes \cite{lin2010construction}. Specifically, operations preserving complete randomness applied to underlying Poisson processes—such as superposition, subsampling, and point transition—yield a new Poisson process and consequently, a new Dirichlet process. 

This approach underscores the versatility of DDP in capturing complex dependencies and temporal dynamics within data, making it a valuable tool across various domains, including time series analysis, dynamic clustering, and sequential modeling.

Constructing Dependent Dirichlet Processes (DDP) from Poisson Processes involves leveraging their fundamental properties through specific operations while maintaining their inherent randomness. The connection between these stochastic processes enables a rigorous mathematical framework for modeling dependencies among random probability measures.

Firstly, the superposition operation combines multiple independent Poisson processes into a single Poisson process. Let \(N_1, N_2, \ldots\) denote independent Poisson processes with rates \(\lambda_1, \lambda_2, \ldots\), respectively. The superposition of these processes, \(N = N_1 + N_2 + \ldots\), yields a new Poisson process with rate \(\lambda = \sum_{i=1}^\infty \lambda_i\). This operation forms the basis for aggregating stochastic events from distinct sources, ensuring that the resulting process retains the stochastic properties of Poisson processes.

Secondly, the subsampling operation involves sampling from a Poisson process at a reduced rate. Given a Poisson process \(N\) with rate \(\lambda\), and a desired lower rate \(r < \lambda\), we introduce a new Poisson process \(M\) with rate \(r\). Sampling from \(N\) at the arrival times of \(M\) yields a subsampled Poisson process with rate \(r\). Mathematically, if \(T_1, T_2, \ldots\) are the arrival times of \(N\), then the arrival times of the subsampled process \(M\) are given by \(T_1', T_2', \ldots\), where \(T_i'\) is the \(i\)th arrival time of \(M\).

Lastly, the point transition operation allows for the transformation of arrival times of a Poisson process according to a given function \(f(t)\). Suppose \(N\) is a Poisson process with rate \(\lambda\), and \(f(t)\) maps each arrival time \(t\) of \(N\) to a new time \(f(t)\). The resulting process, obtained by transforming the arrival times of \(N\) using \(f(t)\), remains a Poisson process but may exhibit a different rate and timing of events.

By employing these operations while preserving the essential randomness of Poisson processes, a new Poisson process and, consequently, a new Dirichlet process can be constructed. This construction provides a robust mathematical foundation for developing Dependent Dirichlet Processes, facilitating the modeling of complex dependencies among random measures in various statistical and machine learning applications \cite{moraffah2018dependent, neiswanger2014dependent, campbell2013dynamic, phung2012conditionally}.

\section{Two-Parameter Poisson Dirichlet Process}

In Section \ref{dp}, we explored the Dirichlet process, a distribution over an infinite-dimensional space that generates clusters in a slow rate regardless of the number of observed data points. However, many real-world phenomena exhibit growth patterns that deviate from this slow rate and instead follow a power-law distribution \cite{newman2005power,pitman1997, pitman2002}. To address this, Pitman et al. introduced a random probability measure known as the Pitman-Yor process, which induces marginal distributions characterized by a two-parameter Chinese restaurant process.

A two-parameter Chinese restaurant process, denoted as CRP$([n],d,\alpha)$, is defined over all partitions with parameters $\alpha>0$ and $d$ ($0 \leq d < 1$ and $\alpha > -d$). This process governs the probability of customers choosing tables (clusters) based on their occupancy and the concentration parameters $\alpha$ and $d$. Specifically, the probability of a customer joining an existing table $\mathcal{C}$ is given by:

\[
\mathbb{P}(\text{Choose table } \mathcal{C}) = \frac{n_{\mathcal{C}} - d}{\alpha + \sum_{\rho} n_{\mathcal{C}}}
\]

where $n_{\mathcal{C}}$ represents the number of customers seated at table $\mathcal{C}$, and the probability of choosing a new table is:

\[
\mathbb{P}(\text{Choose a new table}) = \frac{\alpha + d|\rho|}{\alpha + \sum_{\rho} n_{\mathcal{C}}}
\]

This process is exchangeable, implying the existence of a de Finetti distribution such that the observed data are independent. The de Finetti measure is referred to as the Pitman-Yor process \cite{pitman1997, perman1992size}.
\begin{figure}[th!]
\centering
\includegraphics[width = 14cm, height = 6cm ]{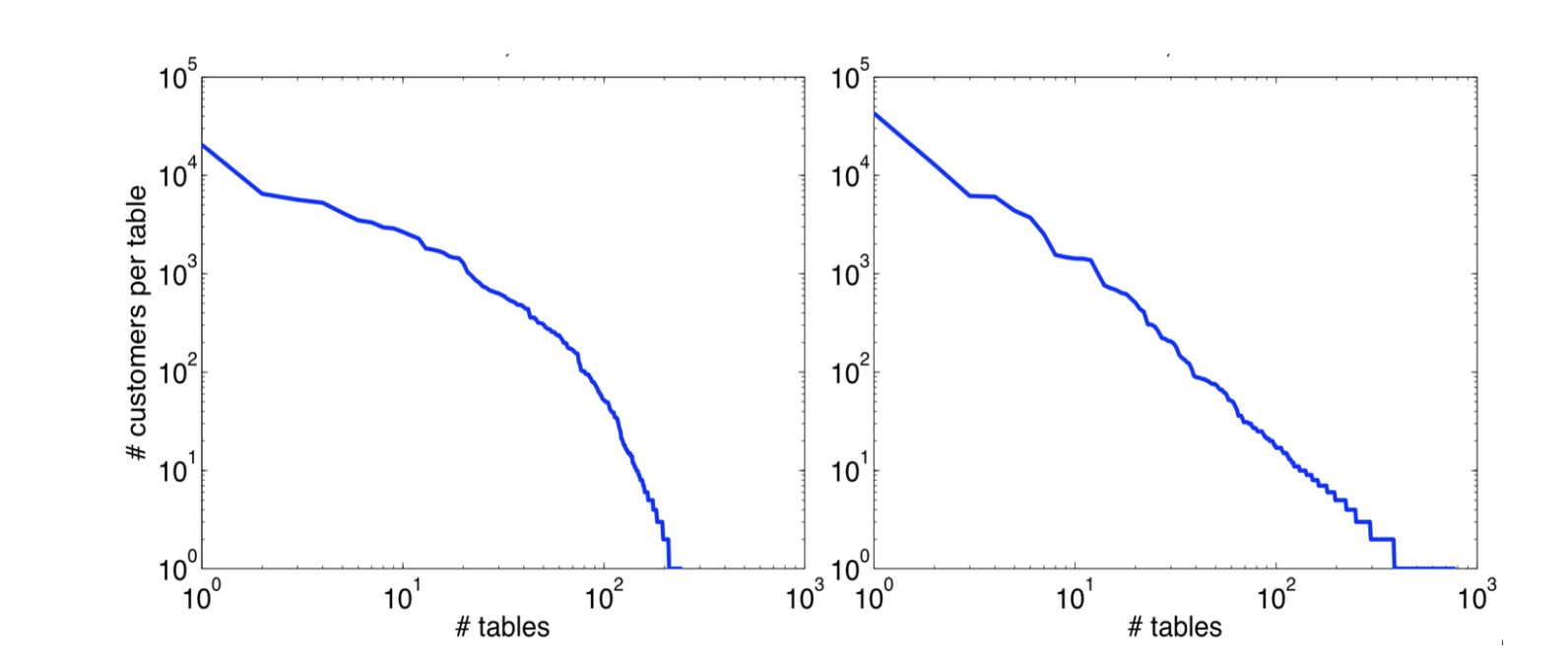} 
\caption{Heap's law for Pitman-Yor process}
\label{fig:py_occ}
\end{figure}

 \begin{figure}[h!]
\centering
\includegraphics[width = 14cm, height = 6cm ]{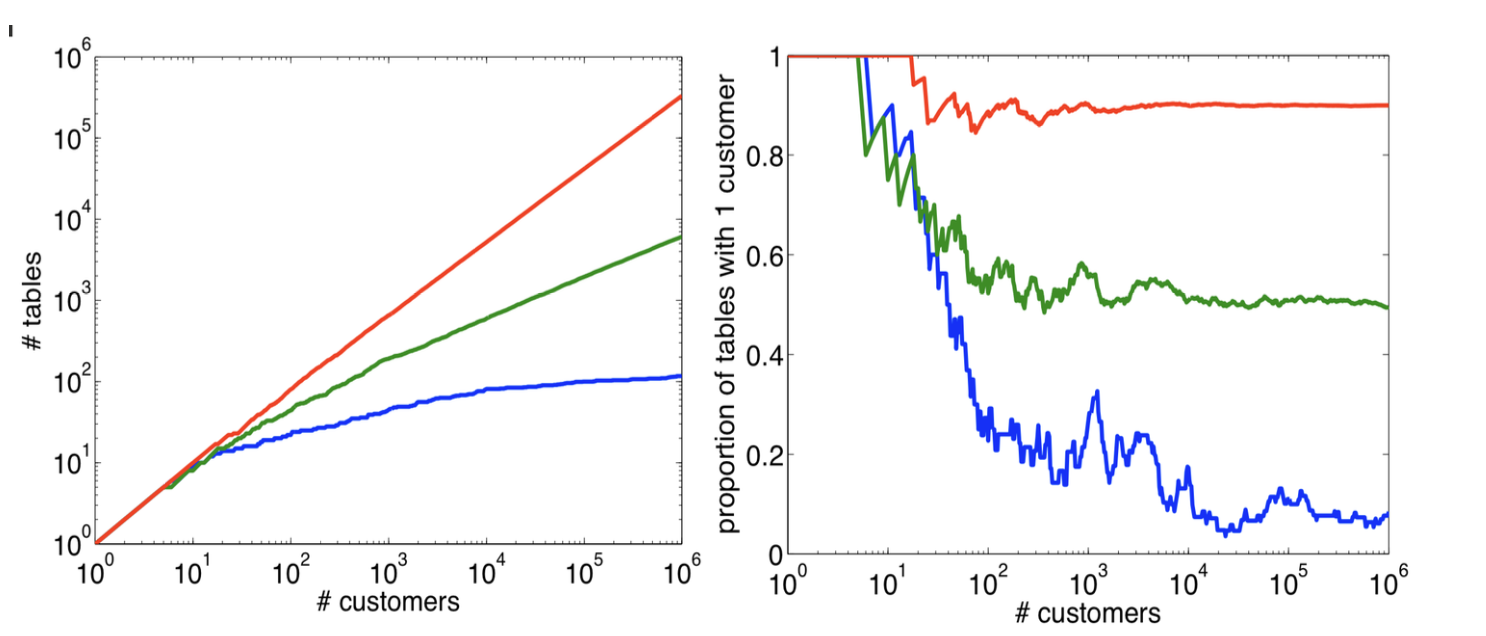} 
\caption{Comparison between Pitman-Yor process and Dirichlet process for $\alpha = 10$ and $d = 0.9$ (red), $d = 0.5$ (green), and $d = 0$ (blue)}
\label{fig:dp-py-heap}
\end{figure}

The Pitman-Yor process is particularly suited for modeling phenomena that exhibit power-law growth patterns, such as natural language text \cite{goldwater2011producing, newman2005power}. By varying the parameters $\alpha$ and $d$, the process demonstrates a willingness to generate more clusters, with tables having more occupants being more likely to grow larger. Conversely, smaller values of $d$ result in more tables with fewer customers. This behavior is illustrated in Figures \ref{fig:py_occ}, \ref{fig:dp-py-heap}, where the Pitman-Yor process exhibits Heap's law, while the Dirichlet process tends to generate fewer tables.

The advent of inferential methods has revolutionized the analysis of high-dimensional data \cite{robert2013monte, andrieu2003, mackay2003}. These methods play a crucial role in scenarios where computing explicit probabilities given parameters is challenging or even infeasible. In this context, two primary inference methods stand out: Monte Carlo methods and variational Bayes. Monte Carlo methods, when coupled with efficient algorithmic designs, have demonstrated remarkable capabilities in providing accurate, exact, and computationally tractable estimates \cite{mackay1998}. On the other hand, variational Bayes offers an attractive approach for approximating intractable integrals or posterior distributions \cite{blei2017variational}. Each of these approaches has its strengths and applications, contributing to the expanding toolkit of statistical inference methods.

\subsection{Stick-Breaking Construction of the Pitman-Yor Process}
The Stick-Breaking Construction of the Pitman-Yor process is a fundamental concept in Bayesian nonparametric statistics, offering a flexible framework for modeling data distributions with unknown and potentially unbounded complexity \cite{james2013stick, james2019stick}. This construction provides a means to define probability distributions over infinite-dimensional spaces, making it particularly useful in scenarios where the number of underlying components or clusters is not fixed a priori. This construction is particularly useful in scenarios where the number of underlying clusters is unknown or potentially unbounded, making it a valuable tool in various applications ranging from natural language processing to machine learning.

At the heart of the Stick-Breaking Construction lies the idea of representing a probability distribution as an infinite mixture model. This model assumes an infinite number of mixture components, each with an associated weight and parameter. The Pitman-Yor process serves as a prior distribution over these mixture weights, allowing for clustering behavior while accounting for uncertainty in the number of clusters.

Let's delve into the mathematical formulation of the Stick-Breaking Construction using the Pitman-Yor process. Suppose we have a set of observations \( \{x_1, x_2, ..., x_n\} \) drawn from an unknown distribution. We aim to model this distribution using a mixture model with an infinite number of components. The weights associated with these components are generated using a stick-breaking process. The stick-breaking process is defined as follows: 

The Stick-Breaking Construction of the Pitman-Yor process begins with a unit-length stick. With each iteration \( k \), a fraction \( \beta_k \) of the remaining stick is broken off, drawn from a Beta distribution characterized by parameters \( \alpha \) and \( \theta \). This broken-off piece is then designated as the weight of the \( k \)-th component in the mixture model under consideration. Through this iterative process, the Stick-Breaking Construction progressively assigns weights to an infinite number of potential mixture components, providing a flexible framework for Bayesian nonparametric modeling with an indefinite number of clusters. Mathematically, the Stick-Breaking Construction can be represented as:

\[ \pi_k = V_k \prod_{j=1}^{k-1} (1 - V_j) \]

Where:
\( \pi_k \) is the weight assigned to the \( k \)-th component, \( V_k \) is a Beta-distributed random variable, \( \alpha \) and \( \theta \) are hyperparameters of the Pitman-Yor process, controlling the concentration and discounting, respectively.

The Pitman-Yor process extends the Dirichlet Process by introducing an additional parameter \( \theta \), which governs the level of clustering. This parameter allows for a power-law behavior in the distribution of cluster sizes, offering greater flexibility in modeling complex datasets \cite{moraffah2019nonparametric, caron2017generalized, Moraffah2019DDPHDP, blunsom2011hierarchical}.

\section{Indian Buffet Process}
The Indian Buffet Process (IBP) is a stochastic process used in Bayesian nonparametric modeling, particularly in the context of sparse feature representations for data. It is closely related to both the Pitman-Yor Process (PYP) and the Beta Process, each offering unique perspectives on infinite-dimensional distributions.

At its core, the IBP is founded upon the Dirichlet Process (DP), which serves as its underlying measure. The DP provides a nonparametric prior distribution over probability measures, allowing for an infinite number of clusters or components. In the context of the IBP, the DP governs the distribution of features across data points, enabling the creation of sparse binary matrices to represent feature presence or absence.

The IBP can be viewed as a discrete counterpart to the PYP, a more general distribution over partitions that extends the capabilities of the DP. While the PYP directly models distributions over partitions, the IBP operates on binary matrices, where each row corresponds to a data point and each column represents a feature. Through a series of binary "dishes," data points are served with features in a probabilistic manner, with the preference for creating new features balanced by the likelihood of reusing existing ones.

In essence, the Indian Buffet Process offers a concrete framework for generating sparse feature representations, leveraging the underlying principles of the Dirichlet Process while drawing inspiration from the broader landscape of Bayesian nonparametrics. By providing a probabilistic mechanism for feature creation and sparsity induction, the IBP serves as a versatile tool in the arsenal of Bayesian machine learning techniques.

The IBP is a probabilistic model used to generate sparse binary matrices, particularly in the context of Bayesian nonparametric modeling. Let \( \mathbf{Z} \) denote the binary matrix generated by the IBP, where each row corresponds to a data point and each column represents a feature. The IBP assumes an infinite-dimensional latent representation, allowing for the creation of an unbounded number of features \cite{griffiths2011indian, ghahramani2005infinite}.

At the heart of the IBP lies a generative process that mimics the metaphor of customers at a buffet. Initially, the first customer (data point) samples features from a Poisson distribution with a specified intensity parameter, representing the expected number of features sampled. Subsequent customers sequentially sample features from the existing set of features, with each feature being chosen with a probability proportional to its prevalence among previous customers.

Mathematically, the IBP can be represented as follows \cite{ghahramani2005infinite}. Let \( K \) be the number of features sampled by the first customer, drawn from a Poisson distribution with intensity parameter \( \lambda \). Subsequent customers sample features according to a probability distribution that depends on the features sampled by previous customers. Let \( m_{nk} \) denote the number of customers who have sampled feature \( k \) up to the \( n \)-th customer. Then, the probability that the \( n \)-th customer samples feature \( k \) is given by:

\[ P(z_{nk} = 1 \,|\, \mathbf{Z}_{-(n,k)}, \alpha) = \frac{m_{nk}}{n} \]

where \( z_{nk} \) is the entry in the \( n \)-th row and \( k \)-th column of \( \mathbf{Z} \), \( \mathbf{Z}_{-(n,k)} \) denotes the binary matrix \( \mathbf{Z} \) with the \( n \)-th row and \( k \)-th column removed, and \( \alpha \) is a parameter controlling the expected number of features. 

The IBP operates on an infinite set of features, denoted by \( K = \infty \), enabling flexible modeling of datasets with potentially countless attributes. Represented by a binary matrix \( \mathbf{Z} \) of size \( N \times K \), where \( N \) signifies the number of data points and \( K \) represents the infinite features, each entry \( z_{nk} \) acts as a binary indicator variable, determining whether data point \( n \) possesses feature \( k \).

The IBP's recursive definition governs how features are sampled for each data point. Initially, for the first data point (\( n = 1 \)), a Poisson-distributed number of features is sampled from an intensity parameter \( \lambda \), yielding a binary matrix with the corresponding number of features. Subsequently, for additional data points (\( n > 1 \)), the sampling process evolves. Each data point decides whether to sample new features or replicate features from preceding data points. This dynamic process results in binary matrices with variable feature counts for each data point, accommodating the varying complexities of the dataset. 


The probability distribution guiding the sampling process is pivotal. The probability of sampling new features for each data point follows a Poisson distribution, parameterized by \( \lambda \), determining the expected number of new features per data point. Additionally, the probability of copying features from prior data points follows a power-law distribution, favoring features with higher counts across data points. This mechanism encourages the emergence of shared features, capturing underlying patterns and structure within the dataset effectively.


 \subsubsection{Properties and Applications}

The Indian Buffet Process (IBP) offers a unique blend of sparsity and flexibility, making it a powerful tool for discovering relevant features within complex datasets. Through its innate mechanism, the IBP induces sparsity in the binary matrix representation, facilitating the automatic identification of pertinent features while effectively managing high-dimensional data. Simultaneously, it exhibits remarkable flexibility, permitting the sampling of an unlimited number of features. This characteristic enables the IBP to capture the potentially infinite intricacies inherent in real-world datasets, providing a robust framework for modeling diverse data structures.

Across various fields encompassing machine learning and statistics, the IBP has garnered widespread adoption for its efficacy in uncovering latent features within datasets of disparate nature. Its applications span a broad spectrum, ranging from image recognition and text analysis to collaborative filtering and biological data analysis \cite{chen2013phylogenetic,pradier2019case, changpinyo2012learning, hu2012modeling}. Leveraging the IBP's ability to uncover underlying patterns, researchers and practitioners have successfully employed it to extract meaningful insights from complex datasets, thus advancing knowledge discovery and decision-making processes across domains.

In practical settings, Bayesian inference techniques play a pivotal role in estimating the parameters of the IBP from observed data. Methods such as variational inference and Markov chain Monte Carlo (MCMC) offer efficient and scalable approaches to infer the intensity parameter \( \lambda \) and the binary matrix \( \mathbf{Z} \). By leveraging these inference methods, researchers can effectively fit IBP models to large-scale datasets, facilitating the extraction of valuable information and insights from increasingly voluminous and complex data sources \cite{gorur2006choice,doshi2009variational,cao2021stick}.

\subsection{Dependent Indian Buffet Process (dIBP)}

The Dependent Indian Buffet Process (dIBP) extends the traditional Indian Buffet Process (IBP) to incorporate dependencies between features, offering a more nuanced framework for modeling complex datasets with structured relationships \cite{pmlr-v9-williamson10a}. This extension enables the modeling of dependencies among features, allowing for richer and more accurate representations of real-world phenomena.

In the dIBP, the binary matrix representation \( \mathbf{Z} \) is augmented to incorporate dependencies between features. Let \( \mathbf{Z} = \{z_{nk}\} \) be the binary matrix of size \( N \times K \), where \( N \) represents the number of data points and \( K \) signifies the number of features. Each entry \( z_{nk} \) indicates whether data point \( n \) possesses feature \( k \).

The dIBP introduces a dependency structure through a dependency matrix \( \mathbf{D} \), where \( d_{kl} \) represents the strength of dependency between features \( k \) and \( l \). This dependency matrix governs the likelihood of jointly activating features, capturing the interrelations between different features within the dataset.

To model dependencies, the probability of activating a feature \( k \) for data point \( n \) is influenced not only by the intensity parameter \( \lambda \) but also by the dependencies between \( k \) and previously activated features. Specifically, the probability \( P(z_{nk} = 1) \) is given by:

\[ P(z_{nk} = 1 | \mathbf{Z}_{\neg n}, \mathbf{D}, \lambda) = \frac{\lambda m_{-k} + \sum_{l=1}^{K} d_{kl} \mathbb{1}(z_{nl} = 1)}{\lambda m_{-k} + \sum_{l=1}^{K} d_{kl} \mathbb{1}(z_{nl} = 1) + \alpha} \]

Where \( \mathbf{Z}_{\neg n} \) denotes the binary matrix \( \mathbf{Z} \) excluding data point \( n \)'s row, \( m_{-k} \) represents the number of data points possessing feature \( k \) excluding data point \( n \), \( \alpha \) is the concentration parameter controlling the overall sparsity of the binary matrix, and  \( \mathbb{1}(\cdot) \) is the indicator function.
This formulation reflects the enhanced probability of activating feature \( k \) for data point \( n \), considering both the intensity parameter \( \lambda \) and the cumulative influence of dependencies with previously activated features.

Bayesian inference methods such as variational inference and Markov chain Monte Carlo (MCMC) are commonly employed to estimate the parameters of the dIBP, including the intensity parameter \( \lambda \), the dependency matrix \( \mathbf{D} \), and the binary matrix \( \mathbf{Z} \), given observed data \cite{pmlr-v9-williamson10a, doshi2009variational}. These methods offer scalable and efficient algorithms for fitting dIBP models to real-world datasets, facilitating the exploration of structured dependencies and the extraction of meaningful insights from complex data structures.

The Dependent Indian Buffet Process presents a powerful framework for modeling dependencies among features, offering a versatile tool for analyzing structured data and uncovering intricate relationships within datasets across various domains, including natural language processing, image analysis, and network modeling. By incorporating dependencies into the feature activation process, the dIBP enhances the expressiveness and fidelity of probabilistic models, enabling more accurate representations of real-world phenomena and supporting richer analyses of complex datasets.

\subsection{Beta Process and Indian Buffet Process} 

The Beta Process is a fundamental distribution in Bayesian nonparametric statistics, often used as a prior distribution over probability measures. Specifically, it generates random probability measures on a continuous domain, typically the unit interval \([0, 1]\). These probability measures represent the distribution of features or attributes within a dataset.

Consider the Indian Buffet Process (IBP), which is a probabilistic model used for generating binary matrices that represent the features of a dataset. In the context of the IBP, each row of the binary matrix corresponds to a data point, and each column represents a feature or attribute. The IBP is characterized by two parameters: the number of features sampled per data point (controlled by \( \lambda \)) and the distribution of features across data points (determined by the base measure \( H \)).

The connection between the Beta Process and the IBP lies in how the Beta Process serves as the underlying mechanism for generating the binary matrices in the IBP. Specifically, the Beta Process generates random probability measures that dictate the distribution of features across data points. Each data point in the IBP samples its features according to these probability measures, resulting in the binary matrix representation of the dataset.

In essence, the Beta Process provides a probabilistic framework for the generative process of the IBP. It governs how features are distributed across data points and influences the sparsity and distribution of features within the binary matrices generated by the IBP. Understanding the Beta Process as the underlying distribution of the IBP illuminates the probabilistic mechanisms driving the generation of binary matrices and elucidates the key characteristics of the IBP, such as its flexibility in handling datasets with varying numbers of features and its ability to capture complex data structures through sparse binary representations.

\subsubsection{Beta Process}
The Beta Process is a stochastic process defined on a continuous domain, typically the unit interval \([0, 1]\). It serves as a prior distribution over an infinite-dimensional space, allowing for the generation of random probability measures. Formally, a Beta Process is denoted as \( \text{BP}(\alpha, H) \), where \( \alpha \) is a positive scalar parameter known as the concentration parameter, and \( H \) is a base measure.

Given a collection of random variables \( \{ G(A) \}_{A \in \mathcal{A}} \), where \( \mathcal{A} \) is a space of subsets, the Beta Process satisfies the following two properties:
\begin{enumerate}
\item Marginal Distribution: For any finite partition \( \{ A_1, A_2, ..., A_n \} \) of the space \( \mathcal{A} \), the joint distribution of \( \{ G(A_1), G(A_2), ..., G(A_n) \} \) follows a Dirichlet distribution with parameters \( \alpha H(A_1), \alpha H(A_2), ..., \alpha H(A_n) \).

\item Independence Property: The random variables \( G(A_1) \) and \( G(A_2) \) are independent if the sets \( A_1 \) and \( A_2 \) are disjoint.
\end{enumerate}
These properties make the Beta Process a versatile tool for modeling random probability measures, providing a flexible framework for Bayesian nonparametric inference.

\subsubsection{Indian Buffet Process as a Beta Process}

The Indian Buffet Process (IBP) is intimately connected to the Beta Process, with the latter serving as its underlying distribution. In the IBP, the Beta Process generates the binary matrices that represent the features of the dataset. Specifically, each row of the binary matrix corresponds to a data point, and each column represents a feature.

The generative process of the Indian Buffet Process (IBP) entails a detailed probabilistic framework that involves the Beta Process as its underlying distribution. 

Initially, we begin with an empty binary matrix \( \mathbf{Z} \) of size \( N \times K \), where \( N \) represents the number of data points and \( K \) signifies the number of features. Each entry \( z_{nk} \) of this matrix denotes whether data point \( n \) possesses feature \( k \). At this starting stage, all entries are initialized to zero, indicating that no features have been activated yet.

For each data point (row \( n \)), the process involves sampling features from a Beta Process prior distribution. This Beta Process, denoted as \( \text{BP}(\alpha, H) \), is characterized by a concentration parameter \( \alpha \) and a base measure \( H \), which represents the distribution of features. Specifically, for each data point \( n \), a random measure \( G_n \) is sampled from the Beta Process, where \( G_n \) represents the distribution of features for data point \( n \). This distribution \( G_n \) is a probability measure defined on the space of features. The concentration parameter \( \alpha \) governs the expected number of features per data point, while the base measure \( H \) captures the prior distribution of features.

Following the determination of the distribution \( G_n \) for each data point \( n \), the subsequent step is to activate features based on the probabilities specified by \( G_n \). Specifically, for each feature \( k \), \( z_{nk} \) is sampled from a Bernoulli distribution with parameter \( G_n(k) \), where \( G_n(k) \) represents the probability of activating feature \( k \) for data point \( n \) as determined by the Beta Process. This activation process is iterated for each data point, resulting in the activation of features across the binary matrix \( \mathbf{Z} \).

By following this generative process, the Indian Buffet Process constructs binary matrices that represent the features of a dataset in a probabilistically principled manner. By leveraging the Beta Process as its underlying distribution, the Indian Buffet Process offers a probabilistic framework for modeling datasets with an indefinite number of features, facilitating flexible and scalable Bayesian nonparametric inference in various applications, including machine learning, natural language processing, and image analysis \cite{chen2011hierarchical,paisley2009nonparametric} . The connection between the Beta Process and the Indian Buffet Process underscores the elegance and versatility of Bayesian nonparametric methods in capturing complex data structures and uncovering latent patterns within datasets.

\subsection{Hierarchical Beta Processes and the Indian Buffet Process}

Hierarchical Beta Processes (HBP) extend the Indian Buffet Process (IBP) by introducing a hierarchical structure that enables the modeling of dependencies among features across different layers \cite{pmlr-v2-thibaux07a}. This hierarchical formulation enriches the IBP framework, allowing for more nuanced representations of complex datasets with hierarchical structures. Understanding the mathematical details of HBP in conjunction with the IBP sheds light on their generative processes and elucidates their key properties.  By introducing dependencies between layers and incorporating feature activations based on Beta Process distributions, it enables the generation of hierarchical representations of features, facilitating more nuanced and interpretable analyses of hierarchical data.

At the core of the Hierarchical Beta Process lies a hierarchical construction that incorporates multiple layers of Beta Processes. Let \( \text{HBP}(\alpha_0, H_0, \gamma) \) denote the Hierarchical Beta Process, where \( \alpha_0 \) is the concentration parameter at the top level, \( H_0 \) represents the base measure at the top level, and \( \gamma \) is the scaling parameter that controls the relationship between adjacent layers. The generative process of the HBP can be described as follows:

Initialization: The generative process of the Hierarchical Beta Process begins with the creation of an empty binary matrix \( \mathbf{Z}^{(0)} \) at the top level. Each entry \( z_{nk}^{(0)} \) in this matrix indicates whether data point \( n \) possesses feature \( k \) at the top level. This initial step sets the foundation for building hierarchical representations of features across multiple layers.

Layer-wise Feature Assignment: At each layer \( l \), ranging from \( l = 1 \) to \( L \), features are sampled from a Beta Process prior distribution. The Beta Process at each layer is characterized by parameters \( \alpha_l \) and \( H_l \), representing the concentration parameter and base measure, respectively. These parameters govern the distribution of features at each layer, shaping the characteristics of the features sampled within the hierarchical structure.

Inter-layer Dependencies: The Hierarchical Beta Process introduces dependencies between adjacent layers using the scaling parameter \( \gamma \). This parameter modulates the influence of features between layers, facilitating the propagation of features across the hierarchical structure. Higher values of \( \gamma \) result in stronger inter-layer dependencies, enabling features to influence activations in subsequent layers, whereas lower values lead to more independent representations across layers.

Feature Activation: Activation of features occurs at each layer based on the sampled probabilities from the corresponding Beta Process distribution. The activation process incorporates dependencies between layers, allowing features activated in one layer to influence activations in subsequent layers. This results in the generation of hierarchical representations of features across the layers of the binary matrix \( \mathbf{Z} \), capturing complex patterns and relationships within the dataset across multiple levels of abstraction.

The generative process of the Hierarchical Beta Process involves probabilistic modeling at each layer, incorporating Beta Process distributions and inter-layer dependencies. Let \( G_n^{(l)} \) denote the random measure at layer \( l \) for data point \( n \), representing the distribution of features at that layer.

At each layer \( l \), the probability of activating feature \( k \) for data point \( n \) is determined by the corresponding random measure \( G_n^{(l)} \). Specifically, \( z_{nk}^{(l)} \) is sampled from a Bernoulli distribution with parameter \( G_n^{(l)}(k) \), where \( G_n^{(l)}(k) \) represents the probability of activating feature \( k \) for data point \( n \) at layer \( l \).

The scaling parameter \( \gamma \) modulates the influence of features between adjacent layers, allowing for the propagation of features across the hierarchy. Higher values of \( \gamma \) result in stronger inter-layer dependencies, whereas lower values lead to more independent representations across layers.

Bayesian inference methods offer a robust framework for estimating the parameters of Hierarchical Beta Processes (HBP) and uncovering the hierarchical structure of features from observed data. Leveraging techniques such as Markov chain Monte Carlo (MCMC) or variational inference enables the exploration of hierarchical dependencies and facilitates the extraction of meaningful insights from complex datasets with hierarchical structures.

MCMC methods, including Gibbs sampling and Metropolis-Hastings sampling, are widely utilized for posterior inference in HBP models. These methods iteratively sample from the posterior distribution over the parameters of the HBP, allowing for the estimation of parameters such as \( \alpha_l \), \( H_l \), and \( \gamma \). By generating a Markov chain of parameter samples, MCMC algorithms provide a flexible and versatile approach to Bayesian inference, accommodating complex dependencies and non-linear relationships within the data.

Variational inference offers an alternative approach to Bayesian inference, aiming to approximate the posterior distribution with a tractable distribution while minimizing the Kullback-Leibler divergence between the approximate and true posterior distributions. In the context of HBP, variational inference optimizes variational parameters to approximate the posterior distribution over the model parameters. This optimization process often involves minimizing a suitable objective function, such as the evidence lower bound (ELBO), to find the best-fitting approximation to the true posterior distribution. Variational inference offers computational efficiency and scalability, making it particularly useful for large-scale datasets and complex models.

In addition to parameter estimation, Bayesian inference methods enable the inference of the hierarchical structure of features within the HBP model. By examining the posterior distributions of model parameters, such as the concentration parameters \( \alpha_l \), and analyzing the relationships between features across different layers, researchers can uncover hierarchical patterns and dependencies within the data. This hierarchical structure provides valuable insights into the organization and representation of features, facilitating a deeper understanding of the underlying data-generating process.

\subsubsection{Applications and Advantages}
The application of Bayesian inference methods in HBP modeling spans various domains, including natural language processing, image analysis, and network modeling. By leveraging hierarchical dependencies, HBP models capture complex relationships and structures within the data, enabling more accurate and interpretable analyses. The flexibility and scalability of Bayesian inference methods allow for the exploration of diverse datasets with hierarchical organization, leading to improved model performance and insights into hierarchical data structures.

\section{Sized-Biased Sampling}

In the context of Bayesian nonparametric models, size-biased sampling refers to a sampling scheme where items are selected with a probability proportional to their ``size" or ``weight." The term ``size" here typically refers to some measure of importance, relevance, or mass associated with each item in the dataset \cite{arratia2019size, scheaffer1972size}.

In Bayesian nonparametric (BNP) models, the utilization of size-biased sampling offers a nuanced approach across several key contexts, each contributing to the refinement and optimization of these models.

First, in the realm of clustering tasks, size-biased sampling emerges as a strategic tool for selecting clusters with a greater abundance of data points. By favoring larger clusters, which often signify more prevalent patterns within the data, BNP clustering algorithms can allocate resources more efficiently. This prioritization enables the automatic allocation of additional attention and resources to clusters that potentially harbor more significant insights, thereby enhancing the clustering process's accuracy and effectiveness.

Second, when grappling with feature selection, BNP methodologies can leverage size-biased sampling to emphasize features with heightened relevance or importance. This tailored sampling strategy ensures that features are sampled in proportion to their significance, facilitating the capture of the most informative facets of the data. Consequently, BNP models can streamline their learning and inference processes, honing in on the most pertinent variables and improving the overall efficacy of the model.

Moreover, within the realm of model evaluation, size-biased sampling takes on a crucial role in constructing evaluation datasets that prioritize instances or scenarios of paramount interest. By emphasizing instances that carry greater significance, this sampling technique ensures that the evaluation process accurately reflects the model's performance on pertinent aspects of the data. As a result, researchers can derive a more precise assessment of the model's capabilities, thereby refining and strengthening the model's applicability across diverse domains and tasks.

In essence, the incorporation of size-biased sampling strategies into BNP models empowers researchers to prioritize items based on their importance or relevance, thereby facilitating more effective learning, inference, and evaluation processes. By harnessing the nuanced capabilities of size-biased sampling, BNP algorithms can navigate complex datasets with greater precision, enhancing their performance and utility across a spectrum of applications and domains.

Size-biased sampling is a concept that finds practical application within the framework of both the Dirichlet Process (DP) and the Pitman-Yor Process (PY) \cite{perman1992size}. This sampling technique involves preferentially selecting items based on their size or prevalence, rather than choosing uniformly from the entire population.

In the Dirichlet Process (DP) mixture model, the random probability measure \( G \) is a pivotal component drawn from the DP prior, characterized by a base measure \( H \) and a concentration parameter \( \alpha \). With a dataset comprising \( N \) data points denoted as \( \{x_1, x_2, \ldots, x_N\} \), the assignment of each data point \( x_i \) to a cluster is determined by sampling from \( G \).

Within the context of DP clustering, size-biased sampling emerges as a methodological approach aimed at selecting clusters with a probability directly proportional to their size. Mathematically, in a partition of the data into clusters \( \{C_1, C_2, \ldots, C_K\} \), the probability of assigning data point \( x_i \) to cluster \( C_k \) is formulated as:
\[ p(C_k | \text{data}) \propto \frac{n_k}{\alpha + N - 1} \]
Here, \( n_k \) represents the count of data points already assigned to cluster \( C_k \), \( \alpha \) signifies the concentration parameter inherent to the DP, and \( N \) denotes the total count of data points in the dataset.

This probabilistic expression encapsulates the essence of size-biased sampling within DP clustering, where the likelihood of assigning a data point to a particular cluster is influenced by the size of the cluster relative to the concentration parameter and the overall dataset size. By favoring larger clusters through this sampling mechanism, the DP clustering algorithm inherently prioritizes clusters that potentially encapsulate more significant patterns within the dataset, thus refining the clustering process and improving its accuracy.

Within Pitman-Yor (PY) mixture models, the Pitman-Yor Process enriches the framework of the Dirichlet Process (DP) by incorporating an additional parameter, \( d \), which modulates the discounting of clusters with smaller sizes. Here, \( G \) denotes the random probability measure drawn from the PY prior, characterized by parameters \( \alpha \) (the concentration parameter) and \( d \) (the discount parameter). Much like in DP clustering, size-biased sampling in PY clustering entails favoring clusters with larger sizes.

Mathematically, the probability of assigning a data point \( x_i \) to cluster \( C_k \) within a PY mixture model is articulated as follows:
\[ p(C_k | \text{data}) \propto \frac{n_k - d}{\alpha + N - 1} + \frac{\alpha}{\alpha + N - 1} \cdot \frac{\Gamma(d)}{\Gamma(d + N)}, \]
where  \( n_k \) represents the count of data points already assigned to cluster \( C_k \),  \( \alpha \) stands for the concentration parameter,  \( d \) signifies the discount parameter, \( N \) denotes the total count of data points in the dataset, and \( \Gamma(\cdot) \) denotes the gamma function.

This formulation encapsulates the intricacies of size-biased sampling within PY clustering, where the likelihood of assigning a data point to a specific cluster hinges on the cluster's size, modulated by both the concentration and discount parameters. The first term in the expression reflects the size-biased sampling component, favoring larger clusters, while the second term introduces additional complexity by considering the discounting effect, which influences the probability based on the size of the cluster and the discount parameter. Through this nuanced probabilistic framework, PY clustering refines the clustering process, enabling the model to capture more nuanced patterns within the data.

In both scenarios, size-biased sampling ensures that clusters with larger sizes (i.e., more data points) are more likely to be sampled during the clustering process, reflecting their greater importance or prevalence in the dataset. This mathematical formulation enables BNP models to capture the prominent patterns or structures in the data and allocate resources more effectively to clusters with greater significance.

\section{Completely Random Measure}

A completely random measure (CRM) is a fundamental concept in probability theory and stochastic processes. Intuitively, it can be understood as a random variable that assigns probabilities to sets in a random manner, embodying uncertainty in a probabilistic system \cite{jordan2010hierarchical}.

To construct a completely random measure intuitively, consider a scenario where you have a set of possible outcomes or events, and you want to assign probabilities to subsets of these outcomes. However, instead of assigning fixed probabilities as in traditional probability theory, you introduce randomness into the process.

One way to construct a CRM is by imagining an infinite collection of dice, each representing a different source of uncertainty. Each die has a potentially different number of sides, and each side corresponds to a different probability distribution over the set of possible outcomes. When you roll these dice, you randomly select one of them, and then roll it to determine the probabilities assigned to subsets of outcomes. The randomness comes from both the selection of the die and the outcome of the roll.

Another intuitive way to think about constructing a CRM is by considering a ``meta-distribution" over probability measures. Imagine you have a bag containing an infinite number of probability distributions. Each time you reach into the bag, you randomly select one of these distributions. Then, based on the selected distribution, you assign probabilities to sets of outcomes according to that distribution. The uncertainty in this process arises from the randomness in selecting which distribution to use.

In both of these intuitive constructions, the key idea is that the probabilities assigned to sets of outcomes are not fixed but rather determined randomly according to some underlying random mechanism. This randomness captures our uncertainty about the true probability distribution governing the system, making completely random measures a powerful tool for modeling uncertainty in various probabilistic contexts.

Formally, let \( (\Omega, \mathcal{F}) \) be a measurable space, where \( \Omega \) is the sample space and \( \mathcal{F} \) is a sigma-algebra representing the set of possible events. A CRM, denoted as \( \mu \), is a random measure on \( (\Omega, \mathcal{F}) \), meaning that for each outcome \( \omega \) in \( \Omega \), \( \mu(\omega, \cdot) \) is a measure on \( (\Omega, \mathcal{F}) \). In other words, for each \( \omega \), \( \mu(\omega, A) \) represents the probability assigned to the event \( A \) given the outcome \( \omega \) \cite{kingman1967completely}.

CRMs are often characterized by their properties. One important property is the measurability of sample paths. That is, for fixed \( \omega \) in \( \Omega \), the function \( A \mapsto \mu(\omega, A) \) should be measurable for every \( A \) in \( \mathcal{F} \). This property ensures that the random measure \( \mu(\omega, \cdot) \) is well-defined for each outcome \( \omega \).

Another key property of CRMs is their independence across disjoint sets. Specifically, if \( A_1, A_2, \ldots, A_n \) are pairwise disjoint sets in \( \mathcal{F} \), then the random variables \( \mu(A_1), \mu(A_2), \ldots, \mu(A_n) \) are independent. This property captures the idea that the probability assigned to one set does not depend on the probabilities assigned to other disjoint sets. In addition, CRMs are often characterized by their distributional properties. For example, CRMs can be specified through their cumulative distribution functions (CDFs) or probability density functions (PDFs). These distributional properties allow us to understand the behavior of CRMs across different outcomes and events.

The aim is to construct a completely random measure (CRM) comprised solely of non-fixed atoms within the space \( \Theta \). This endeavor necessitates the utilization of an augmented space, specifically the product space \( \Xi = \Theta \times \mathbb{R}^+ \), where \( \mathbb{R}^+ \) denotes the set of nonnegative real numbers. By introducing a Poisson random measure \( N \) over this product space \( \Theta \times \mathbb{R}^+ \), we proceed to formulate the CRM on \( \Theta \) as a function dependent on \( N \).

Let's assume that the mean measure of \( N \) over \( \Theta \times \mathbb{R}^+ \) can be represented in the product form as:
\[ \mu = G_0 \otimes \nu, \]
where \( G_0 \) signifies a diffuse probability measure on \( \Theta \), and \( \nu \) denotes a diffuse measure on \( \mathbb{R}^+ \). This representation clarifies that for sets \( A \subseteq \Theta \) and \( E \subseteq \mathbb{R}^+ \), the measure \( \mu \) assigns probabilities in accordance with \( G_0(A) \nu(E) \). Here, \( G_0 \) is termed the base (probability) measure on \( \Theta \), while \( \nu \) constitutes the Lévy measure of the CRM.

Furthermore, employing the notation \( \mu(d\phi, dw) = G_0(d\phi) \nu(dw) \), we envision taking integrals with respect to the measure \( \mu \). Given a Poisson random measure \( N \) on \( \Theta \times \mathbb{R}^+ \), let \({(\phi_k, w_k)} \) denote the set of atoms in a specific realization of \( N \). This realization can be represented as:
\[ N = \sum_k  \delta_{\phi_k, w_k}, \]
where \( \delta_{\phi_k, w_k} \) is an atom located at \({(\phi_k, w_k)} \)  in the product space. Based on these atoms, we construct a measure \( G \) on \( \Theta \) as a weighted sum of atoms, where \( \phi_k \) specifies the location of the \( k \)-th atom and \( w_k \) represents its mass:
\[ G = \sum_k w_k \delta_{\phi_k}. \]
It's noteworthy that the range of \( k \) isn't specified as finite or infinite, yet an infinite range can indeed be achieved through the selection of \( \nu \).

Clearly, \( G \) emerges as a random measure, with \( G(A) \) representing the total sum of masses for those atoms \( \phi_k \) falling within \( A \):
\[ G(A) = \sum_{k: \phi_k \in A} w_k. \]
Moreover, as the underlying Poisson random measure \( N \) is completely random, \( G \) also inherits complete randomness. Specifically, if \( A_1, A_2, \ldots, A_K \) are disjoint subsets of \( \Theta \), then the sets \( \{A_j \times \mathbb{R}^+\} \) are also disjoint. Consequently, the restrictions of \( N \) to these subsets are mutually independent, thereby rendering the masses \( G(A_1), G(A_2), \ldots, G(A_K) \) mutually independent as well \cite{kingman1967completely}.

Once we have constructed a completely random measure (CRM) \( G \) from the Poisson random measure \( N \), we have the flexibility to augment \( G \) with additional elements. Firstly, we can introduce a set of fixed atoms to \( G \), where the locations of these fixed atoms are predetermined prior to generating \( G \). These fixed atoms possess mutually independent masses that are also independent of \( N \). This augmentation allows for the incorporation of specific points of interest into the CRM without altering its fundamental random structure.

Furthermore, we have the option to include a deterministic measure into this augmented CRM. Remarkably, the resulting object remains a completely random measure. This property underscores the versatility and generality of CRMs, as they can accommodate deterministic components alongside their inherently random nature.

In particular, this approach fully characterizes CRMs, demonstrating that all CRMs can be systematically constructed as a sum of three distinct components:
\begin{enumerate}
\item[1.] The original CRM formed from the Poisson random measure \( N \).
\item[2.] A set of fixed atoms with predetermined locations and independent masses, which are also independent of \( N \).
\item[3.] An optional deterministic measure component.
\end{enumerate}

By allowing for the incorporation of fixed atoms, deterministic measures, or both, this construction framework offers a comprehensive understanding of the structure and composition of CRMs. It highlights the rich interplay between randomness and determinism in these probabilistic objects, elucidating their utility and versatility in various theoretical and applied contexts.

To exemplify the overarching construction of CRMs elucidated in the preceding section let's delve deeper into the specifics of each process:

The Gamma Process (GP) offers a foundational framework within Bayesian nonparametric modeling, particularly for scenarios involving continuous distributions. In the context of the GP, the mean measure \( \mu \) is characterized by a base measure \( G_0(d\phi) \) on the positive real line \( \mathbb{R}^+ \), augmented by parameters \( \alpha \) and \( \beta \) governing the shape and rate of a Gamma distribution, respectively. This representation enables the construction of flexible probability distributions that adapt to the observed data.

More specifically, within the GP formulation, the mean measure \( \mu \) is expressed as \( \mu(d\phi, dw) = G_0(d\phi) \alpha w^{-1} e^{-\beta w} dw \), where \( \phi \) represents the location parameter on \( \mathbb{R}^+ \), and \( w \) denotes the scale parameter. The term \( G_0(d\phi) \) encapsulates the base measure, dictating the distribution of \( \phi \) across the positive real line. Meanwhile, the hyperparameters \( \alpha \) and \( \beta \) finely tune the shape and rate of the underlying Gamma distribution, respectively.

The expression \( w^{-1} e^{-\beta w} \) delineates the probability density function of the Gamma distribution with parameters \( \alpha \) and \( \beta \). This formulation ensures that the mean measure \( \mu \) adheres to the Gamma distribution's probability density characteristics across different locations \( \phi \). Through this intricate interplay of parameters and distributions, the GP engenders a versatile probabilistic framework capable of capturing complex data structures and patterns in continuous domains.

The Beta Process (BP) constitutes another fundamental Bayesian nonparametric model, particularly pertinent for scenarios involving distributions defined on bounded intervals, such as the unit interval \( [0, 1] \). Within the framework of the BP, the mean measure \( \mu \) is characterized by a base measure \( G_0(d\theta) \) on the unit interval \( [0, 1] \), along with a scale parameter \( w \) governing the shape of the distribution.

In the formulation of the BP, the mean measure \( \mu \) can be explicitly defined as \( \mu(d\theta, dw) = G_0(d\theta) w^{-1}(1-\theta)^{w-1} d\theta \), where \( \theta \) denotes the location parameter within the unit interval \( [0, 1] \). The scale parameter \( w \) further modulates the distribution's shape, influencing the probability density across different locations \( \theta \).

The term \( G_0(d\theta) \) signifies the base measure on the unit interval \( [0, 1] \), dictating the distribution of the location parameter \( \theta \). Meanwhile, the expression \( w^{-1}(1-\theta)^{w-1} \) encapsulates the probability density function of the Beta distribution, ensuring that the mean measure \( \mu \) conforms to the characteristics of a Beta distribution at each location \( \theta \). The Beta Process furnishes a versatile probabilistic framework adept at capturing intricate data structures and patterns within bounded intervals. This formulation enables the Beta Process to model a diverse array of phenomena characterized by distributions confined within the unit interval \( [0, 1] \), offering valuable insights and inference capabilities across various domains.

The Dirichlet Process (DP) stands as a cornerstone within Bayesian nonparametric modeling, especially in scenarios involving distributions defined on the probability simplex \( \Delta \) in \( \mathbb{R}^d \). Within the framework of the DP, the mean measure \( \mu \) is characterized by a base measure \( G_0(d\boldsymbol{\theta}) \) on the probability simplex, along with a scale parameter \( w \) influencing the distribution's shape.

In the context of the DP, the mean measure \( \mu \) can be succinctly expressed as \( \mu(d\boldsymbol{\theta}, dw) = G_0(d\boldsymbol{\theta}) w^{-1} e^{-w} d\boldsymbol{\theta} \), where \( \boldsymbol{\theta} \) represents a probability vector within the probability simplex \( \Delta \) in \( \mathbb{R}^d \). The scale parameter \( w \) further modulates the distribution's shape, influencing the probability density across different probability vectors \( \boldsymbol{\theta} \).

The term \( G_0(d\boldsymbol{\theta}) \) denotes the base measure on the probability simplex, delineating the distribution of the probability vectors \( \boldsymbol{\theta} \). Meanwhile, the expression \( w^{-1} e^{-w} \) encapsulates the probability density function of the Dirichlet distribution, ensuring that the mean measure \( \mu \) conforms to the characteristics of a Dirichlet distribution at each probability vector \( \boldsymbol{\theta} \). By intricately intertwining parameters and distributions, the Dirichlet Process furnishes a versatile probabilistic framework adept at capturing intricate data structures and patterns within the probability simplex \( \Delta \) in \( \mathbb{R}^d \). This formulation enables the Dirichlet Process to model a diverse array of phenomena characterized by distributions confined within the probability simplex, offering valuable insights and inference capabilities across various domains.

\subsubsection{L\'evy-Khinchin Formula }

The L\'evy-Khinchin formula is a fundamental result in probability theory that provides a characterization of the characteristic function of a probability distribution \cite{kallenberg1997foundations}. In the context of Bayesian nonparametrics (BNP), the L\'evy-Khinchin formula is particularly relevant in understanding the stochastic processes underlying BNP models.

L\'evy-Khinchin formula states that the characteristic function \( \phi(u) \) of a probability distribution \( P \) can be expressed as:
\[ \phi(u) = \exp\left[ i \mu u - \frac{1}{2} \sigma^2 u^2 + \int_{\mathbb{R}} \left( e^{iux} - 1 - iux \mathbb{I}_{|x| < 1} \right) \Pi(dx) \right] \]
where \( \mu \) is the mean of the distribution, \( \sigma^2 \) is the variance of the distribution, and \( \Pi \) is the L\'evy measure, which captures the jump component of the distribution.

In the context of BNP, this formula is often used in the study of stochastic processes such as the Dirichlet Process (DP) and its variants. The DP, for example, can be viewed as a distribution over distributions, where the Lévy measure plays a crucial role in characterizing the jump behavior of the process.

To understand the L\'evy--Khinchin formula in the context of BNP, we initially focus on computing the expected value of \( G(A) \), where \( G(A) \) follows a gamma distribution \( \text{Gamma}(a, b) \), with given shape parameter \( a \) and inverse scale parameter \( b \). This expectation \( E[G(A)] \) should equate to \( a/b \). We begin by expressing \( G(A) \) in a more convenient form, where \( G \) is defined in terms of a Poisson random measure \( N \). We rewrite \( G(A) \) as an integral over the discrete measure \( N \):
\[ G(A) = \sum_{k} w_k \delta_{\phi_k}(A) = \int_{\mathbb{R}^+} w \mathbb{I}_{\{\phi \in A\}} N(d\phi, dw) \]

Here, \( \mathbb{I}_A(\phi) \) denotes an indicator function that equals one if \( \phi \) is in \( A \) and zero otherwise. By simplifying the notation further, defining \( \xi = (\phi, w) \), we rewrite the integral as:
\[ G(A) = \int_{\Xi} f(\xi) N(d\xi) \]

This integral represents the integration under the Poisson random measure \( N \) of a function on the space \( \Xi = \Theta \times \mathbb{R}^+ \). To compute the expectation of such an integral, we first focus on computing the expectation of integrals of step functions and gradually extend to more complex functions.

By considering step functions of the form \( f(\xi) = \sum_{j \in J} c_j \mathbb{I}_{C_j}(\xi) \), where \( J \) is a finite index set and \( C_j \) are non-overlapping sets, we compute:
\[ E\left[\int_{\Xi} f(\xi) N(d\xi)\right] = \sum_{j \in J} c_j \mu(C_j) = \int \mu(d\xi) f(\xi) \]

Here, \( \mu \) represents the mean measure. This computation extends to general positive functions \( f(\xi) \) using the monotone convergence theorem, allowing us to compute the first moments of a Poisson random measure. This result is known as Campbell's theorem or the first-moment formula for Poisson random measures and summarized as: 

\begin{theorem}[Campbell's theorem] For general positive functions, we can express the expected value of the integral \( \int_{\Xi} f(\xi) N(d\xi) \) as the integral of \( f(\xi) \) with respect to the mean measure \( \mu \), which is given by:
\[ E\left[\int_{\Xi} f(\xi) N(d\xi)\right] = \int_{\Xi} f(\xi) \mu(d\xi). \]
\end{theorem}

Moving forward, utilizing the Monotone Convergence Theorem for BNP, we can establish that for general positive functions \( f(\xi) \), the Laplace transform can be evaluated as:
\[ \mathbb{E}\left[e^{-t \int f(\xi) N(d\xi)}\right] = \exp\left\{- \int (1 - e^{-tf(\xi)}) \mu(d\xi)\right\} \]
This significant result, known as the Lévy-Khinchin formula, underscores the pivotal role of the mean measure in computing probabilities associated with Poisson random measures.

To utilize this framework in Bayesian nonparametrics (BNP), we can apply these concepts to model the underlying stochastic processes in BNP models, such as the Dirichlet Process (DP) or its variants. By understanding the expected behavior of these stochastic processes, we gain insights into the probabilistic properties of BNP models, facilitating inference and model development in complex probabilistic settings.

\subsection{Normalized Completely Random Measure}
A Normalized Completely Random Measure (NCRM) is a type of random measure that is defined on a measurable space and exhibits normalization properties. Similar to a Completely Random Measure (CRM), an NCRM is characterized by its distributional properties and its behavior under integration.

Formally, let \( (\Omega, \mathcal{F}) \) be a measurable space, where \( \Omega \) is the sample space and \( \mathcal{F} \) is a sigma-algebra representing the set of possible events. An NCRM \( \mu \) on \( (\Omega, \mathcal{F}) \) is a random measure with the following properties:
\begin{enumerate}
\item Normalization: The Normalized Completely Random Measure \( \mu \) is defined such that its total mass over the sample space \( \Omega \) is finite almost surely. This can be mathematically expressed as:
   \(\mu(\Omega) < \infty \) which implies that with probability one, the NCRM assigns a finite total mass to the entire sample space.

\item Randomness: Similar to a Completely Random Measure (CRM), an NCRM is a stochastic object whose properties are determined by a probability distribution. Consequently, the behavior of the NCRM can vary across different realizations, reflecting inherent uncertainty and variability.

\item Measure-Theoretic Properties: An NCRM exhibits measure-theoretic properties akin to those of any random measure. These include sigma-additivity and countable additivity, ensuring that the NCRM behaves in a manner consistent with a measure. This characteristic enables the integration of functions over the sample space, facilitating rigorous mathematical treatment and analysis.
\end{enumerate} 
It is worth noting that the normalization condition ensures that the NCRM assigns finite probabilities to events, making it suitable for probabilistic modeling and inference. This property distinguishes an NCRM from a CRM, where the total mass may be infinite.

\subsubsection{Stick-Breaking Representation of Normalized Random Measure}

The Stick-Breaking Representation is a method used to construct Normalized Completely Random Measures (NCRMs) from a sequence of beta-distributed random variables. This representation is a foundational concept in Bayesian nonparametric modeling, particularly in the context of Dirichlet Processes and their extensions.

Let's denote \( \{ V_k \}_{k=1}^{\infty} \) as a sequence of independent beta-distributed random variables, where \( V_k \sim \text{Beta}(1, \alpha) \) with \( \alpha > 0 \). These beta-distributed random variables are used to generate the weights for the atoms of the NCRM. The Stick-Breaking Representation defines the NCRM \( \mu \) as follows:
 \begin{enumerate}
\item Construction of Atom Weights: Each atom weight \( w_k \) of the NCRM is obtained by multiplying the sequence of beta-distributed random variables with a decreasing sequence of stick lengths. Specifically, we define \( w_k = V_k \prod_{j=1}^{k-1} (1 - V_j) \) for \( k = 1, 2, \ldots \).

\item Normalization: To ensure that the total mass of the NCRM is finite, we normalize the weights by dividing each weight by the sum of all weights. Mathematically, the normalized weight \( \tilde{w}_k \) is given by \( \tilde{w}_k = \frac{w_k}{\sum_{j=1}^{\infty} w_j} \).

\item  Construction of the NCRM: Finally, the NCRM \( \mu \) is defined as a random measure supported on the atoms generated by the stick-breaking process, with atom weights given by \( \tilde{w}_k \).
\end{enumerate}
The Stick-Breaking Representation of the NCRM can be summarized as:
\[ \mu = \sum_{k=1}^{\infty} \tilde{w}_k \delta_{\theta_k} \]
where \( \theta_k \) represents the location of the \( k \)-th atom and \( \tilde{w}_k \) denotes the normalized weight associated with the \( k \)-th atom.

This representation allows for the construction of NCRMs with finite total mass, providing a flexible framework for modeling complex probability distributions in Bayesian nonparametric inference.
\subsubsection{Dependent Normalized Completely Random Measures}

Dependent Normalized Completely Random Measures (DNCRM) extend the concept of Normalized Completely Random Measures (NCRM) to scenarios where there is dependence among the random measures \cite{lijoi2014bayesian}. The Dependent Normalized Completely Random Measure (DNCRM) \( \mu \) is defined by a sequence of random measures \( \{\mu_i\}_{i=1}^{\infty} \) on \( \mathcal{X} \), each characterized by its own distribution. The key features of a DNCRM are its dependence structure, normalization properties, and joint distribution.

The dependence structure in Dependent Normalized Completely Random Measures (DNCRM) indicates that each random measure \( \mu_i \) is influenced by the distributions of the preceding measures \( \mu_1, \mu_2, \ldots, \mu_{i-1} \). This dependency is expressed through conditional distributions, denoted as \( P(\mu_i | \mu_1, \mu_2, \ldots, \mu_{i-1}) \), which mathematically represent how the distribution of \( \mu_i \) is affected by its predecessors.

To ensure that the total mass over the measurable space \( \mathcal{X} \) is finite almost surely, each individual random measure \( \mu_i \) is normalized. Mathematically, this is stated as \( \mu_i(\mathcal{X}) < \infty \) for every \( i \), where \( \mathcal{X} \) represents the measurable space. This normalization condition guarantees that the DNCRM remains well-defined and avoids infinite mass issues.

The joint distribution of the sequence \( \{\mu_i\}_{i=1}^{\infty} \) comprehensively captures the interplay among the random measures. It elucidates how the distributions of the individual measures interact and evolve throughout the entire sequence. This joint distribution, being multivariate, characterizes the entire DNCRM \( \mu \), and its formulation is contingent upon the conditional distributions and dependencies among the individual \( \mu_i \).

\subsection{Poisson-Kingman Processes}

Poisson-Kingman Processes (PKP) are a class of Bayesian nonparametric (BNP) models used to model random probability measures \cite{kingman1992poisson, pitman2003poisson}. They are particularly useful in applications involving clustering, species sampling, and other scenarios where an unknown number of categories or clusters is expected. PKP generalizes the popular Dirichlet Process (DP) and Pitman-Yor Process (PY) models.

Let \( \Theta \) be a measurable space representing the parameter space, and \( G_0 \) be a base probability measure on \( \Theta \). The PKP is defined by a random probability measure \( G \) that follows a Poisson-Kingman distribution, denoted as \( G \sim \text{PK}(\gamma, G_0) \), where \( \gamma \) is the Poisson-Kingman parameter.

The Poisson-Kingman parameter \( \gamma \) controls the behavior of the PKP and determines the expected number of clusters or categories. Higher values of \( \gamma \) result in a larger number of clusters, while smaller values of \( \gamma \) lead to fewer clusters.

The PKP can be characterized by its stick-breaking construction, similar to the DP and PY processes. Let \( \{V_k\}_{k=1}^{\infty} \) be a sequence of independent random variables, each following a beta distribution \( \text{Beta}(1, \gamma) \). The stick-breaking weights \( \{W_k\}_{k=1}^{\infty} \) are obtained as \( W_k = V_k \prod_{j=1}^{k-1} (1 - V_j) \).

The PKP assigns a weight \( W_k \) to each cluster, where \( k \) indexes the clusters. The locations of the clusters are drawn independently from the base measure \( G_0 \). Specifically, the probability measure \( G \) can be represented as:
\[ G = \sum_{k=1}^{\infty} W_k \delta_{\theta_k} \]
where \( \delta_{\theta_k} \) is a Dirac delta function centered at \( \theta_k \), representing the location of the \( k \)-th cluster.

\subsubsection{Poisson-Kingman partition}
Poisson-Kingman processes play a crucial role in Bayesian nonparametric (BNP) modeling, particularly concerning partition structures and clustering. Central to this framework is the concept of Poisson-Kingman partitions, which represent random partitions of a set into an unknown number of subsets \cite{pitman2003poisson}.

In this setting, a Poisson-Kingman partition \( \Pi \) describes the random partition of a set \( \mathcal{X} \). Its probability distribution follows a Poisson-Kingman distribution, where the probability mass function assigns probabilities to partitions based on their sizes. This distribution is typically parameterized by a concentration parameter, governing the likelihood of observing partitions of different sizes.

Complementing Poisson-Kingman partitions are Poisson-Kingman processes, stochastic processes used to model the generation of such partitions. A Poisson-Kingman process \( \mathcal{PK} \) defines a probability distribution over partitions of \( \mathcal{X} \). This distribution is determined by a probability measure that specifies how partitions of varying sizes are generated. Common choices for this measure include the Pitman-Yor process or the normalized generalized gamma process, each offering different properties and flexibility in modeling.

The relation between Poisson-Kingman partitions and processes is tightly intertwined. While the Poisson-Kingman process dictates the mechanism by which partitions are generated, the resulting partitions adhere to the distribution specified by the Poisson-Kingman partition. In essence, the process generates partitions, and the resulting partitions conform to the distribution characterized by the partition, facilitating probabilistic inference over partition structures.

In Bayesian nonparametric modeling, Poisson-Kingman partitions and processes serve as versatile tools for capturing uncertainty and variability in partition structures. They enable flexible modeling of random partitions, making them invaluable in applications such as clustering, community detection, and feature allocation. Through the interplay between Poisson-Kingman partitions and processes, BNP methods offer a powerful framework for exploring and understanding complex data structures.

\subsection{Gibbs-Type Exchangeable Random Partition}

Gibbs-Type Exchangeable Random Partition (GTERP) models are pivotal in Bayesian nonparametric statistics, offering a versatile approach for modeling random partitions of a set \cite{lijoi2010models, ghosal2017fundamentals, de2011, lijoi2008investigating}. These models are instrumental in capturing the uncertainty inherent in partition structures and find widespread application across diverse domains such as clustering, topic modeling, and community detection.

A GTERP defines a probability distribution over partitions of a set \( \mathcal{X} \), denoted by \( \Pi \). The distribution of \( \Pi \) is characterized by a probability measure, typically specified using a hierarchical Bayesian model. One key characteristic of GTERP models is their exchangeability property, signifying that the distribution of partitions remains invariant to the ordering of elements in \( \mathcal{X} \). This exchangeability ensures that the model treats all elements equally, irrespective of their ordering. The exchangeability property of GTERP models is formalized as \( P(\Pi = \pi) = P(\Pi = \pi \circ \sigma) \) for any permutation \( \sigma \) of the elements in the set \( \mathcal{X} \), where \( \pi \) and \( \pi \circ \sigma \) represent partitions corresponding to the original and permuted sets, respectively. This property asserts that the probability of observing a particular partition remains unchanged under permutations of the elements in \( \mathcal{X} \), ensuring that the model treats all elements equally regardless of their ordering.

Moreover, GTERP models exhibit a Gibbs-type property, implying that the conditional distribution of a partition given relevant information follows a Gibbs distribution. This property facilitates the use of efficient inference algorithms, such as Markov chain Monte Carlo (MCMC) methods, for posterior sampling. In Gibbs-Type Exchangeable Random Partition (GTERP) models, the conditional distribution of a partition given the parameters \( \theta \) follows a Gibbs distribution, represented as \( P(\Pi \mid \theta) = \frac{1}{Z(\theta)} \exp(-\mathcal{H}(\Pi \mid \theta)) \), where \( \mathcal{H}(\Pi \mid \theta) \) is the Hamiltonian function and \( Z(\theta) \) is the normalization constant. The Hamiltonian function encapsulates the energy or cost associated with a particular partition given the parameters \( \theta \). This Gibbs property enables efficient probabilistic inference by providing a principled framework for sampling partitions based on their energy levels, ensuring that partitions with lower energy are more likely to be sampled. In practice, parameters of the GTERP model, including concentration parameters and hyperparameters governing the partition distribution, are often assigned prior distributions within a hierarchical Bayesian framework.

Inference in Gibbs-Type Exchangeable Random Partition (GTERP) models revolves around estimating the posterior distribution of the parameters \( \theta \) given observed data, a task critical for understanding the underlying partition structure. While the hierarchical Bayesian framework provides a principled approach for specifying prior distributions and likelihood functions, obtaining the posterior distribution often requires advanced computational techniques. Markov chain Monte Carlo (MCMC) methods emerge as powerful tools for sampling from the posterior distribution, enabling robust inference in complex models like GTERP.

In the realm of MCMC techniques for posterior sampling in GTERP models, Gibbs sampling, Metropolis-Hastings, and score-based MCMC algorithms are notable choices. Gibbs sampling operates by iteratively sampling from the conditional distributions of each parameter given the others, leading to a Markov chain whose stationary distribution converges to the posterior distribution. Meanwhile, the Metropolis-Hastings algorithm offers flexibility by allowing proposals from arbitrary distributions, though it may exhibit slower convergence rates. Nevertheless, both methods are adept at navigating the high-dimensional parameter space inherent in GTERP models, thereby enabling accurate estimation of the posterior distribution. Score-based MCMC algorithms, on the other hand, utilize gradient information or other derivative-based measures to guide the proposal distribution, often resulting in more efficient exploration of the parameter space and improved convergence properties.

In the domain of MCMC techniques for posterior sampling in GTERP models, Gibbs sampling and Metropolis-Hastings are prominent options. Gibbs sampling iteratively samples from the conditional distributions of each parameter given the others, driving a Markov chain to converge to the posterior distribution. Conversely, the Metropolis-Hastings algorithm permits proposals from arbitrary distributions, though it may demonstrate slower convergence. Nevertheless, both methods effectively navigate the high-dimensional parameter space intrinsic to GTERP models, facilitating precise estimation of the posterior distribution. 

Moreover, sophisticated Markov Chain Monte Carlo (MCMC) techniques like Hamiltonian Monte Carlo (HMC), Langevin MCMC, and Sequential Monte Carlo (SMC) methods have risen in significance for inference tasks involving intricate Bayesian models. These score-based MCMC methods employ gradient information or other derivative-based metrics to guide the proposal distribution, thereby enabling more effective traversal of the parameter space and enhanced convergence characteristics. HMC utilizes Hamiltonian dynamics to formulate proposals that comprehensively explore the parameter space, resulting in improved mixing and accelerated convergence rates. Similarly, SMC methods present a systematic means to approximate the posterior distribution through a series of weighted samples, offering superior precision and scalability compared to conventional MCMC methodologies.

\section{Bayesian Nonparamterics and Deep Learning Alternative}

Bayesian nonparametrics (BNP) offers a compelling alternative to traditional deep learning methods, providing flexible and interpretable frameworks for modeling complex data structures. While deep learning has achieved remarkable success in various domains, BNP presents unique advantages and applications that complement and, in some cases, surpass deep learning approaches. Here are a few applications that BNP excels over deep neural networks.
 
 \subsubsection{ Handling Uncertainty}

Bayesian nonparametric (BNP) methods stand out in their ability to naturally capture and quantify uncertainty within model predictions, a feature that holds significant value in domains where uncertainty estimation is critical. Tasks such as medical diagnosis, autonomous driving, and financial forecasting require not only accurate predictions but also an understanding of the associated uncertainty to make informed decisions.

In medical diagnosis, for instance, uncertainty estimation is paramount due to the high stakes involved. Doctors not only need to know the predicted diagnosis but also the confidence level associated with it. BNP models excel in this regard by providing probabilistic predictions that convey the model's uncertainty. Instead of just providing a single diagnosis, BNP models offer a probability distribution over possible diagnoses, allowing medical professionals to gauge the reliability of the predictions and make decisions accordingly, such as recommending further tests or treatments.

Similarly, in autonomous driving, uncertainty estimation plays a crucial role in ensuring safe navigation. BNP models can provide probabilistic predictions about the positions and movements of other vehicles, pedestrians, and obstacles. This information allows autonomous vehicles to assess the reliability of their perception systems and make appropriate decisions, such as adjusting speed or trajectory based on the level of uncertainty associated with detected objects.

In financial forecasting, uncertainty estimation is essential for risk management and investment decision-making. BNP models can provide probabilistic forecasts for asset prices, stock returns, or market trends, along with measures of uncertainty. Investors and financial analysts can use this information to assess the risk associated with different investment strategies and make more informed decisions about portfolio management and asset allocation.

In contrast, traditional deep learning models typically provide point estimates, which lack information about the model's uncertainty. While deep learning models may achieve high accuracy on certain tasks, they often fail to quantify uncertainty, leaving decision-makers in the dark about the reliability of predictions. BNP models offer a principled framework for capturing uncertainty and providing probabilistic predictions, enabling decision-makers to make more informed and risk-aware decisions in critical domains.

 \subsubsection{Scalability and Robustness}

Deep learning models have garnered widespread attention for their remarkable performance on various tasks, but their reliance on large amounts of labeled data and substantial computational resources presents significant challenges, especially in resource-constrained environments. The need for extensive data and computational power can be prohibitive, particularly in settings where data acquisition is costly or limited.

Deep learning models typically require large labeled datasets to learn complex patterns and relationships in the data effectively. However, collecting and annotating such datasets can be time-consuming, expensive, and sometimes impractical. Moreover, training deep neural networks often involves computationally intensive tasks, requiring powerful hardware accelerators such as GPUs or TPUs and substantial computational resources. This reliance on extensive data and computational resources can pose challenges for deployment in real-world scenarios, particularly in settings with limited access to such resources.

In contrast, BNP methods offer scalability and robustness by automatically adapting model complexity to the available data, making them well-suited for resource-constrained environments. BNP models are characterized by their ability to grow in complexity as more data becomes available, without the need for explicit specification of the model size or structure. This inherent flexibility allows BNP models to effectively learn from small datasets and adapt to noisy or incomplete data settings, making them particularly useful in scenarios where data is scarce or noisy.

Furthermore, BNP methods offer robust performance in settings where data quality may vary or where the underlying data distribution is unknown or complex. By capturing uncertainty and incorporating prior knowledge into the modeling process, BNP models can effectively handle uncertainties and ambiguities in the data, leading to more reliable and interpretable results.

The scalability and robustness of BNP methods make them attractive alternatives to deep learning models, particularly in resource-constrained environments where data availability and computational resources are limited. By leveraging the inherent flexibility and adaptability of BNP methods, researchers and practitioners can develop models that are well-suited for real-world applications, even in challenging settings with limited resources.

 \subsubsection{ Model Interpretability}

BNP models offer a unique advantage over deep learning models in terms of interpretability. Unlike deep neural networks, which are often treated as black boxes due to their complex architectures and high-dimensional parameter spaces, BNP models provide interpretable representations of data and learned patterns, enabling users to gain insights into the underlying data structures and relationships.

One key aspect of BNP models is their transparent probabilistic framework, which allows for a clear understanding of model outputs and the uncertainty associated with predictions. BNP models are built on probabilistic principles, which means that they explicitly model uncertainty in the data and the parameters of the model. This transparency allows users to interpret model outputs in a meaningful way and understand the factors driving predictions.

In contrast, deep learning models, particularly deep neural networks, often lack transparency and interpretability. The high-dimensional nature of neural network architectures, coupled with the large number of parameters, can make it challenging to understand how the model arrives at its predictions. This lack of interpretability can be problematic, especially in critical applications where understanding model decisions is essential for trust and accountability.

BNP models offer interpretable representations of data and learned patterns through their probabilistic framework. By explicitly modeling uncertainty and providing probabilistic predictions, BNP models enable users to assess the reliability of predictions and make informed decisions. Additionally, BNP models can uncover underlying data structures and relationships, providing valuable insights into the data generating process.

The interpretability of BNP models makes them well-suited for applications where understanding model decisions and interpreting predictions are critical. By providing transparent probabilistic frameworks, BNP models empower users to gain insights into their data and make informed decisions based on reliable and interpretable model outputs.

 \subsubsection{Incorporating Prior Knowledge}

BNP methods offer a powerful framework for incorporating prior knowledge and domain expertise into the model design process, allowing practitioners to seamlessly integrate existing knowledge with observed data to enhance model performance. This capability is particularly valuable in scenarios where domain knowledge is abundant or where certain structural properties of the data are known a priori.

One of the key advantages of BNP methods is their flexibility in specifying prior distributions. Unlike traditional deep learning models, which typically rely solely on data-driven learning, BNP models allow practitioners to encode prior beliefs, assumptions, and constraints into the model through the choice of priors. These priors can be tailored to reflect domain-specific knowledge and capture known patterns or relationships in the data.

By incorporating prior knowledge into the modeling process, BNP methods enable practitioners to leverage existing information and guide the learning process more effectively. This integration of prior knowledge can lead to more efficient learning and better generalization performance, especially in scenarios where the available data is limited or noisy. Additionally, by allowing for the specification of flexible priors, BNP methods can adapt to different modeling tasks and data settings, providing a versatile framework for model development.

Furthermore, BNP methods facilitate the exploration of complex data structures and relationships by allowing practitioners to specify hierarchical priors and incorporate dependencies between model parameters. This hierarchical modeling approach enables the modeling of latent structures and hierarchies within the data, leading to more interpretable and robust models.

In contrast, deep learning models often lack mechanisms for explicitly incorporating prior knowledge and domain expertise. While deep neural networks excel at learning complex patterns from large-scale data, they may struggle to generalize effectively in settings where domain knowledge is crucial or where the data distribution deviates from the training distribution. By contrast, BNP methods offer a principled approach to integrating prior knowledge with data-driven learning, providing a more holistic and flexible framework for model development.

Overall, the ability of BNP methods to incorporate prior knowledge and domain expertise into the modeling process makes them a valuable alternative to deep learning models, particularly in scenarios where interpretability, generalization, and the integration of existing knowledge are critical considerations. By leveraging the flexibility and adaptability of BNP methods, practitioners can develop models that better capture the underlying structure of the data and provide more reliable and interpretable predictions.

 \subsubsection{Flexibility in Model Complexity}

BNP models provide a flexible and adaptive framework for modeling complex data structures, offering several advantages over traditional deep learning architectures. One key advantage is their ability to automatically discover latent patterns in the data and adaptively allocate model resources, leading to improved generalization performance and mitigating the risk of overfitting.

Deep learning architectures often require manual tuning of hyperparameters and network architectures to achieve optimal performance. This process can be time-consuming, labor-intensive, and prone to suboptimal choices, especially in scenarios with limited domain expertise or understanding of the data. In contrast, BNP methods alleviate the need for manual tuning by automatically adjusting model complexity to fit the data.

BNP models accomplish this through their inherent flexibility in model specification. By allowing for the specification of flexible priors and incorporating hierarchical structures, BNP methods can adaptively adjust model complexity to capture the underlying patterns and relationships in the data. This adaptability enables BNP models to automatically discover latent patterns and allocate resources more efficiently, leading to improved model generalization and robustness.

Moreover, BNP methods inherently incorporate uncertainty into model predictions, providing a natural mechanism for regularization and mitigating the risk of overfitting. Instead of providing deterministic point estimates, BNP models offer probabilistic predictions, which inherently account for uncertainty in the data and the model parameters. This uncertainty regularization helps prevent the model from fitting noise in the data excessively, leading to more robust and generalizable models.

Furthermore, BNP methods offer a principled approach to model selection and complexity control. Instead of relying on ad-hoc methods or heuristic approaches for selecting model architectures and hyperparameters, BNP models use Bayesian inference to automatically infer the appropriate model complexity from the data. This Bayesian approach ensures that the model complexity is well-aligned with the available data, leading to improved generalization performance and more reliable predictions.

BNP models offer a flexible and adaptive framework for modeling complex data structures, enabling the automatic discovery of latent patterns and the adaptive allocation of model resources. By alleviating the need for manual tuning and providing a principled approach to complexity control, BNP methods offer significant advantages over traditional deep learning architectures, leading to improved model generalization and robustness in a wide range of applications.

While deep learning has revolutionized many fields with its remarkable performance on large-scale tasks, Bayesian nonparametrics offers a complementary approach that addresses key challenges in uncertainty quantification, scalability, interpretability, integration of prior knowledge, and flexibility in model complexity. As such, BNP serves as a powerful alternative to deep learning, particularly in domains where uncertainty estimation, interpretability, and robustness are critical considerations. By leveraging the strengths of BNP methods alongside deep learning techniques, researchers and practitioners can develop more comprehensive and effective solutions for a wide range of applications.

\section{Future Directions}
The future of Bayesian nonparametric (BNP) models holds promising directions for their integration with or as alternatives to deep neural networks (DNNs). Here are some potential future directions that in my view could provide promising models in conjunction with or as alternatives to deep neural networks, with exciting opportunities for advancing the state-of-the-art in machine learning and addressing key challenges in model interpretability, uncertainty quantification, robustness, and scalability. By leveraging the complementary strengths of BNP methods and deep learning architectures, researchers can develop more versatile, interpretable, and reliable models for a wide range of applications.

\textbf{Hybrid Models:} One promising avenue entails crafting hybrid models that amalgamate the strengths of both Bayesian nonparametric (BNP) methods and deep neural networks (DNNs). These hybrid architectures could exploit the flexibility and interpretability inherent in BNP methods for specific components of the model, while tapping into the powerful representation learning capabilities offered by DNNs for other aspects. By seamlessly integrating these two paradigms, such hybrid models could offer a potent framework for tackling intricate tasks that demand both adaptability and effective representation learning. These tasks may span various domains, including natural language processing, computer vision, and reinforcement learning, where the ability to combine the robustness of BNP methods with the expressive power of DNNs could unlock new frontiers in model performance and understanding.

\textbf{Interpretable Deep Learning:} As the demand for interpretable and transparent machine learning models continues to rise, there is a burgeoning interest in enhancing the interpretability of deep learning architectures. Deep neural networks (DNNs) often operate as black boxes, making it challenging to understand the factors influencing their predictions and the underlying data structures they capture. In contrast, Bayesian nonparametric (BNP) methods offer a principled approach to interpretability, providing transparent probabilistic frameworks that allow users to glean insights into model outputs and understand the uncertainty associated with predictions.

Integrating BNP principles into deep learning architectures holds significant promise for improving model interpretability while maintaining high performance. By incorporating BNP techniques, such as probabilistic modeling and uncertainty quantification, into DNNs, it becomes possible to elucidate the latent patterns and relationships encoded within the data. These integrated models could provide not only accurate predictions but also meaningful insights into the underlying data structures and the model's decision-making process.

One approach to integrating BNP principles into deep learning architectures involves augmenting traditional neural network layers with Bayesian counterparts. For example, Bayesian neural networks (BNNs) replace deterministic weights and biases with probability distributions, allowing for uncertainty estimation in model predictions. By incorporating BNP-inspired layers into deep learning architectures, practitioners can harness the interpretability and uncertainty quantification capabilities of BNP methods while leveraging the expressive power of DNNs.

Furthermore, integrating BNP principles into deep learning frameworks can enhance model transparency by providing interpretable representations of learned features and patterns. BNP methods inherently capture uncertainty in model predictions, enabling users to assess the reliability of predictions and understand the factors driving model decisions. By leveraging BNP techniques, such as hierarchical modeling and nonparametric priors, deep learning architectures can offer more transparent and interpretable representations of complex data, fostering trust and understanding among users.

\textbf{Uncertainty Quantification:} The capacity of Bayesian nonparametric (BNP) models to inherently capture uncertainty in predictions represents a crucial asset across numerous applications. This feature becomes particularly indispensable in domains where robust and reliable predictions are paramount, such as autonomous driving and medical diagnosis. By integrating uncertainty quantification techniques from BNP into deep learning architectures, there lies a significant opportunity to enhance the reliability of predictions and bolster safety measures in these safety-critical domains.

BNP models excel at providing probabilistic predictions that encapsulate uncertainty, enabling decision-makers to gauge the confidence level associated with each prediction. In autonomous driving, for instance, uncertainty quantification is essential for ensuring safe navigation, especially in complex and dynamic environments. By incorporating BNP-inspired uncertainty quantification techniques into deep learning models deployed in autonomous vehicles, it becomes possible to provide more robust and reliable predictions of vehicle trajectories, object detection, and collision avoidance, thereby enhancing overall safety and trustworthiness.

Similarly, in medical diagnosis, uncertainty quantification plays a pivotal role in guiding clinical decision-making and treatment planning. BNP models offer a principled framework for quantifying uncertainty in diagnostic predictions, taking into account factors such as data variability and model uncertainty. By integrating BNP-inspired uncertainty quantification techniques into deep learning models used for medical image analysis, clinicians can obtain more reliable predictions of disease diagnosis and prognosis, along with confidence intervals that reflect the uncertainty associated with each prediction. This enables more informed and confident decision-making, leading to improved patient care and outcomes.

Moreover, integrating uncertainty quantification techniques from BNP into deep learning architectures can help mitigate the risks associated with model uncertainty and robustness in safety-critical applications. By providing probabilistic predictions and uncertainty estimates, these integrated models offer a transparent and interpretable framework for assessing the reliability of predictions and making informed decisions under uncertainty. This enhances the resilience of deep learning models to unforeseen scenarios and ensures safer operation in safety-critical domains.

\textbf{Small Data and Few-shot Learning:} BNP models possess a remarkable ability to excel in learning from small datasets and dynamically adjusting model complexity to accommodate the available data. Looking ahead, future research could be directed towards developing BNP-inspired methodologies tailored for few-shot learning and meta-learning scenarios, where only a scant amount of labeled data is accessible. Such endeavors hold the potential to revolutionize the field by enabling more efficient learning from limited data and enhancing generalization performance.

In few-shot learning, the challenge lies in training models to make accurate predictions when provided with only a handful of labeled examples per class. BNP-inspired approaches could offer a solution by leveraging the inherent flexibility of BNP models to adaptively adjust model complexity based on the available data. By employing nonparametric priors and hierarchical structures, these models can effectively capture the underlying data distribution and make reliable predictions even with limited training samples. Additionally, BNP-inspired methodologies could incorporate uncertainty quantification techniques to provide confidence estimates for predictions, thereby enabling more informed decision-making in few-shot learning scenarios.

Similarly, in meta-learning tasks, where models are trained on a diverse range of tasks and expected to generalize to unseen tasks, BNP-inspired approaches could play a pivotal role. By leveraging the adaptive nature of BNP models, these methodologies can efficiently learn task-specific representations and adapt to new tasks with minimal data. Incorporating hierarchical Bayesian frameworks into meta-learning architectures could facilitate the sharing of knowledge across tasks while allowing for task-specific variations, leading to more robust and transferable models.

Furthermore, BNP-inspired approaches for few-shot learning and meta-learning could benefit from advancements in scalable inference algorithms and optimization techniques. By developing efficient inference procedures tailored for BNP models, researchers can overcome computational challenges associated with modeling complex data structures and handling large-scale datasets. These advancements could enable the practical deployment of BNP-inspired methodologies in real-world applications, where computational efficiency is paramount.

\textbf{Robustness and Adversarial Defense:} BNP models exhibit a remarkable capability to mitigate overfitting and adapt model complexity according to the data at hand. This inherent robustness against overfitting stems from the flexibility of BNP models to automatically adjust their complexity based on the observed data distribution. Looking forward, there is considerable potential in integrating BNP principles into deep learning architectures to bolster robustness and resilience to adversarial attacks.

Deep learning models are susceptible to adversarial attacks, where small, imperceptible perturbations to input data can lead to erroneous predictions. By incorporating BNP principles into deep learning architectures, it becomes possible to enhance robustness against such attacks. BNP-inspired approaches offer a principled framework for capturing uncertainty and modeling data distributions more effectively, thereby improving the generalization performance of deep learning models in real-world settings.

One approach to incorporating BNP principles into deep learning architectures involves leveraging Bayesian neural networks (BNNs), which replace deterministic weights and biases with probability distributions. By representing model parameters as probability distributions, BNNs can capture uncertainty in predictions and provide more reliable estimates of model uncertainty. This uncertainty quantification enables BNNs to identify areas of input space where predictions are less certain, thereby improving the robustness of deep learning models to adversarial attacks.

Furthermore, BNP-inspired approaches can enhance robustness by modeling data distributions more effectively. Traditional deep learning models often make strong assumptions about the underlying data distribution, which may not hold true in real-world scenarios. BNP methods, on the other hand, offer a flexible and adaptive framework for modeling complex data distributions without relying on strong parametric assumptions. By capturing uncertainty and modeling data distributions more accurately, BNP-inspired approaches can improve the robustness of deep learning models to adversarial attacks and improve their performance in real-world settings.

Moreover, BNP-inspired techniques can enhance model robustness by incorporating techniques such as ensemble learning and model averaging. By training multiple models with different initializations or subsets of the training data and combining their predictions, BNP-inspired approaches can reduce the impact of adversarial attacks and improve the robustness of deep learning models to various perturbations in the input data.

\textbf{Scalability and Efficiency:} While BNP methods boast scalability and robustness, ongoing research aims to enhance their scalability, especially for handling large-scale datasets and high-dimensional data. Future advancements in scalable inference algorithms and optimization techniques hold the potential to facilitate the practical deployment of BNP models in large-scale applications, either autonomously or in conjunction with deep learning approaches.

Scalability is a crucial consideration for deploying BNP models in real-world settings, particularly as datasets grow larger and more complex. While BNP methods offer flexibility and adaptability, their computational complexity can pose challenges for handling massive datasets. To address this, researchers are actively developing scalable inference algorithms that can efficiently process large-scale data while maintaining the accuracy and reliability of BNP models. These algorithms leverage techniques such as stochastic optimization, variational inference, and parallel computing to scale BNP methods to datasets of unprecedented size.

Moreover, advancements in optimization techniques play a vital role in improving the scalability of BNP models. By developing efficient optimization algorithms tailored for BNP frameworks, researchers can overcome computational bottlenecks and expedite the model training process. These optimization techniques may include novel optimization algorithms, adaptive learning rate schedules, and regularization strategies designed to enhance convergence and stability in large-scale settings.

Furthermore, the integration of BNP models with deep learning approaches offers a promising avenue for enhancing scalability and addressing the challenges associated with large-scale data. By leveraging the scalability and robustness of BNP methods in conjunction with the expressive power of deep learning architectures, researchers can develop hybrid models capable of handling massive datasets while providing interpretable and uncertainty-aware predictions.

\section{Conclusion}
While deep learning has undoubtedly made significant strides in various domains of artificial intelligence, it's crucial to acknowledge that it may not always be the most suitable option. Bayesian nonparametrics provides a flexible and principled alternative that shines in situations where uncertainty, model adaptability, and data efficiency are critical factors. By leveraging the unique strengths of both approaches, researchers and practitioners can explore new avenues and tackle a wider array of challenges in AI and machine learning, ultimately leading to more robust and versatile solutions.

\bibliographystyle{ba}
\bibliography{references.bib}

\end{document}